\documentclass[10pt]{article} % For LaTeX2e
\usepackage[preprint]{tmlr}
\usepackage{amsmath,amsfonts,bm}

\def\eqref#1{equation~\ref{#1}}
\def\1{\bm{1}}

\DeclareMathAlphabet{\mathsfit}{\encodingdefault}{\sfdefault}{m}{sl}
\SetMathAlphabet{\mathsfit}{bold}{\encodingdefault}{\sfdefault}{bx}{n}

\usepackage[dvipsnames]{xcolor} 
\usepackage{multirow}
\usepackage{hyperref}
\hypersetup{
    colorlinks=true,
    citecolor=Blue, 
    linkcolor=Blue, 
    urlcolor=Blue
}
\usepackage{url}
\usepackage[table]{xcolor}
\usepackage{colortbl}
\usepackage{microtype}
\usepackage{graphicx}
\usepackage{subcaption}
\usepackage{booktabs}
\usepackage{hyperref}

\usepackage{amsmath}
\usepackage{amssymb}
\usepackage{mathtools}
\usepackage{amsthm}
\usepackage[capitalize,noabbrev]{cleveref}
\usepackage{multirow} 
\usepackage{rotating} 

\definecolor{darkgreen}{rgb}{0.0, 0.4, 0.0}
\newcommand{\improv}[1]{\scriptsize{\textcolor{darkgreen}{(#1)}}}
\definecolor{firebrick}{rgb}{0.698, 0.133, 0.133}
\newcommand{\worse}[1]{\scriptsize{\textcolor{firebrick}{(#1)}}}

\title{Interleaved Noise Injection Improves Clean, Corrupted, and OOD Performance}
\author{\name Matt L. Wiemann \email matt.sampson@princeton.edu \\
      \addr Princeton University
      \AND
      \name Peter Melchior \email peter.melchior@princeton.edu \\
      \addr Princeton University
      \AND
      \name Andrew K. Saydjari \email andrew.saydjari@princeton.edu\\
      \addr Princeton University}

\def\month{MM}  % Insert correct month for camera-ready version
\def\year{2026} % Insert correct year for camera-ready version
\def\openreview{\url{https://openreview.net/forum?id=XXXX}} % Insert correct link to OpenReview for camera-ready version

\usepackage{arydshln}

\definecolor{lightgray}{gray}{0.7}
\arrayrulecolor{black}
\theoremstyle{plain}

\theoremstyle{definition}

\theoremstyle{remark}

\usepackage[textsize=tiny]{todonotes}

\begin{document}
\maketitle

\begin{abstract}
  Noise injection is a well-known technique in stochastic optimization. We report its surprising effectiveness with an interleaved (on-off-on-off...) rather than the usual monotonic decay schedule.
  We present a theoretical analysis of noise injection, which confirms that corruption by impulse (salt-and-pepper) noise approximates a Jacobian regularization, whereas Gaussian noise acts as a curvature penalty.
  This regularization behavior has been invoked to explain why noise injection increases model robustness. 
  But the interleaved nature of our proposed schedule produces superior results even for the optimization objective: mixing phases of noisy data permits the optimizer to escape local minima and increase exploration without the risk of catastrophically forgetting the important features from the clean data.
  To stabilize this training scheme against the rapid changes of the loss when switching between clean and noisy data, we introduce a gradient-norm stabilization technique that scales noisy updates based on clean gradient magnitudes. 
  We compare this method with other common augmentation methods and find substantial improvements in corruption tolerance and robustness to real-world distribution shifts on CIFAR-100-C, ImageNet-C, and ImageNet-R datasets for ResNet and ViT architectures, with the best results being achieved when stacking our method on top of other augmentations. 
  Through saliency and attention maps we show that the effect of interleaved noise injection stems from penalizing the failure modes encouraged by the inductive bias of the models: impulse noise works against the locality bias of convolutional (ResNet) architectures, and Gaussian noise reduces the tendency of attention-based models to pick up large-scale spurious features.
  Interleaved noise injection is therefore an effective tool to improve the test performance on clean, noisy, and out-of-distribution data at essentially zero computational cost. 
\end{abstract}

\section{Introduction}
Stochastic gradient descent (SGD) and its variants are widely used as optimization algorithms for complex, non-convex functions, which often contain many under-performing local minima. One well-studied aspect of SGD, which is theorized to contribute to the method's success, is the presence of noise \citep{hochreiter1994simplifying, hochreiter1997flat, hardt2016train, wu2020direction, xie2021positive, battash2024revisiting}, typically through the mini-batch process, whereby a noisy estimate of the true gradient signal is taken to update the weights of the model. This introduction of stochasticity helps to escape local minima and saddle points that otherwise may trap an optimizer.

Consequently, it has been recognized that deliberate noise injection, typically the addition of Gaussian noise to the data, further improves robustness and generalization \citep{matsuoka1992noise,grandvalet1997noise,zur2009noise,he2019parametric, xie2021positive,spiking_noise,xiao2024noise}. Adding Gaussian noise to the weights is employed for sharpness-aware-minimization \citep{foret2020sharpness} and, after training, to create a diverse ensemble of models \citep{gan2026neural}. However, for optimization methods with noise injection one has to decide on a noise schedule, and each choice comes with downsides. A monotonic decaying schedule restricts the optimizers capacity to escape suboptimal minima late in training, whereas ``curriculum'' (\citet{curriculum_bengio}, which starts with no noise and then switches to high noise) or ``anti-curriculum'' (which does the opposite) schedules either lose their regularization power or underfit the clean data, respectively: these schedules risk forgetting what they fit to during the early training phase. But this need not be the case. Rather than decaying the noise, or turning it on or off completely, we propose an interleaved schedule with several sharp transitions between clean epochs and corrupted epochs. By evaluating both additive (Gaussian) and non-additive (impulse) noise perturbations within this schedule, we demonstrate that interleaved noise injections improves clean, corrupt, and out-of-distribution (OOD) performance. While the latter two performance gains are not surprising, the improvement on clean data (compared to training only on clean data) deserves further investigation because it suggests using noise as a tool to improve not only the robustness but the outright training objective. The remainder of this paper presents a theoretical derivation of the effects of injected noise, empirical evidence for our claim of superior performance, and an investigation into the interplay between neural network architecture and the most effective type of noise corruption.

% the first figure 
  \begin{figure}[t]
      \centering
      \includegraphics[width=\linewidth]{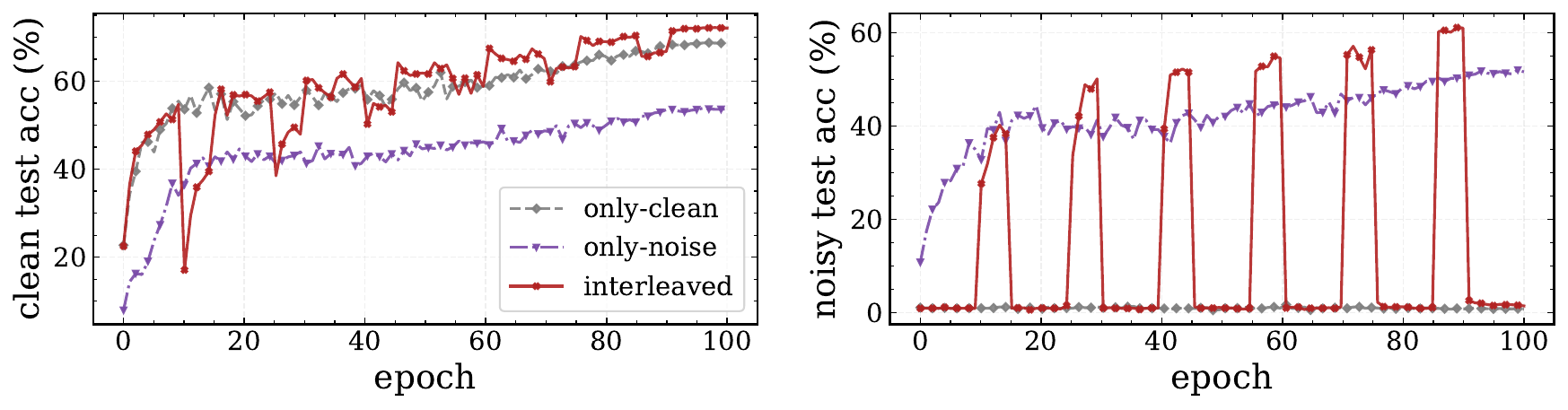}
      \caption{Test accuracy during training for models trained on clean data only (gray), noisy data only (purple) or with our method of interleaved noise-injection epochs ($P=10,L=5$, solid red line). Each curve stems from a single optimization run for WideResNet, trained from scratch on CIFAR-100. The left panel shows the accuracy on held-out clean data; the right panel shows the same for held-out noisy data.}
      \label{fig:clean_vs_injected}
  \end{figure}

\subsection*{Contributions}

\begin{itemize}
    \item \textbf{Interleaved noise curriculum:} We introduce an interleaved noise curriculum for stochastic optimization, which outperforms standard curriculum and anti-curriculum methods on clean, corrupted, and out-of-distribution classification tasks. 
    \item    \textbf{Theoretical formulation of impulse noise regularization:} We present a second-order Taylor expansion to demonstrate how intermittently injecting noise acts as a regularizer. In particular, we derive that impulse noise approximates directional Jacobian regularization. 
    \item \textbf{Empirical improvements and orthogonality to existing regularizers:}
    We test the interleaved curriculum against other corruption and robustness regularization techniques on CIFAR-100 and ImageNet-1k datasets on WideResNet, ResNet50 and ViT architectures. Interleaved noise injection improves clean accuracy, corrupted accuracy, and robustness across all tests. We show that layering the interleaved noise schedule on top of existing regularization methods further improves performance across all three domains.
    \item \textbf{Gradient-norm stabilization:} A primary challenge in optimization with noise injection is the volatility of the gradient signal and its magnitude during noisy epochs. We present a dynamic scaling method to constrain the magnitude of the optimizer updates to prevent vanishing and/or exploding gradients for noise-injection optimization. 
    \item \textbf{Architecture specific noise preference:}  We provide empirical evidence showing convolutional-based networks benefit more from impulse noise, whereas attention-based models perform better under Gaussian noise. The specific noise distribution reduces overfitting for the specific inductive biases of each  architecture. We show that injection with the appropriate noise distribution for a specific architecture always results in model improvements.
\end{itemize}

\section{Related work}

%\paragraph{Simulated annealing} Simulated annealing is an optimization method originating from the field of statistical physics, with a goal of finding deeper minima on optimization surfaces containing many local sub-minima \citep{SA}. The general update equation in simulated annealing for some arbitrary loss function $\mathcal{L}$ can be cast as an SDE;
%\begin{equation}
%    \label{eq:annealing}
%    d\theta_t = - \nabla \mathcal{L}(\theta_t)dt + \sqrt{2T_t} dW,
%\end{equation}
%where $\theta$ is the model parameters, $W_t$ is a standard Wiener process, and $T_t$ represents the time-dependent temperature schedule. Here, the standard accept/reject step $\text{exp}\{-\boldsymbol{\delta}/T_t\}$ is replaced by $\sqrt{2T_t} dW_t$ term in \autoref{eq:annealing}. By systematically injecting noise (or taking a random step) one increases the ability to escape local minima. 

\paragraph{Simulated (and parallel) tempering}
Simulated tempering involves simultaneously performing stochastic optimization on some loss surface over a range of fixed temperatures \citep{marinari1992simulated, earl2005parallel}. These methods aim to reduce the occurrence of optimizers becoming stuck in a sub-optimal local minima. As the action of adding noise to samples is equivalent to increasing the temperature, we see our work in this tradition, with the main contribution coming from temporal modulation of the noise.

\paragraph{Noise-injection for regularization:} \citet{franzke_annealing} explored injecting noise into datasets for MCMC algorithms and found this method is able to speed up convergence when using monotonically decaying noise curricula, akin to simulated annealing. \citet{xie2021positive} explored the effect of simulating stochastic gradient noise, finding that using their ``positive-negative momentum'' method improved generalization performance over a variety of image classification tasks. The role of noise as a regularization method is also explored in \citet{wu2020direction} who empirically find that SGD noise due to mini-batch sampling is Hessian-dependent and pushes the optimizer towards flatter minima. This result hints at noise structure, as well as magnitude, being important factors affecting regularization effectiveness. \citet{levi2022noise} introduce Noise Injection Node Regularization (NINR), which adds additional noise-injecting nodes into the architecture of NNs. This acts to adaptively penalize the loss sharpness, which leads to flatter solutions but introduces additional training times and architecture alterations.

\paragraph{Sharpness-aware-minimization}
Stochastic noise can be injected directly into the model weights during optimization. Classical weight noise approaches add Gaussian perturbations to the parameters to encourage the optimizer to settle in flat minima, thereby improving generalization. More recently, this concept has been formalized by Sharpness-Aware Minimization (SAM) \citep{foret2020sharpness}, which explicitly seeks parameters whose neighborhood has, ideally, uniformly low loss. While weight noise and SAM operate on the parameter space to implicitly or explicitly minimize the trace of the Hessian with respect to the weights, we will show that impulse noise  acts equivalently but operates entirely on the input space. This allows our method to approximate directional Jacobian regularization and minimize input-loss sharpness, specifically guarding against corruptions and out-of-distribution input shifts rather than parameter perturbations.

%\paragraph{Randomized smoothing}  \citet{cohen2019certified} demonstrated that by augmenting the input data with Gaussian noise during training, and smoothing the classifier's prediction at inference time, one can derive tight certification bounds against adversarial perturbations. While standard randomized smoothing typically relies on additive Gaussian noise ($\mathbb{E}[\delta]=0$) to vanish first-order Taylor expansion terms, our work explores an impulse noise regime. Here, the noise distribution is non-additive and the expected perturbation is non-zero, effectively retaining the first-order gradient terms to approximate directional Jacobian regularization rather than just curvature smoothing. See \autoref{sec:schedule} for more details.

\paragraph{Curriculum learning:} \citet{curriculum_bengio} first introduced the concept of curriculum learning for the field of machine learning, typically proposing a curriculum for data samples then moves monotonically from easy (clean) to hard (noisy or complex). Conversely, anti-curriculum and golden-sample methods work to expose the optimizer to the most difficult samples first. Interleaving is the process of alternating between distinct tasks or concepts during learning, and has been consistently shown to improve long-term retention and generalization compared to blocked, continuous practice in student learning \citep{taylor2010effects, rohrer2015interleaved, firth2021systematic}. We draw from these lines of work and propose an interleaved curriculum, which aims to improve the overall performance of trained models.

\section{Methods}
\label{sec:methods}
In this section, we formalize our interleaved noise curriculum and show its regularization effects on the optimization landscape through a second-order Taylor expansion. We then introduce our gradient-norm stabilization routine which allows for stable training even at very high noise levels. We end by describing the range of experiments we perform in this study.

\subsection{Interleaved noise regularization}
\label{sec:schedule}

The noise injection schedule $N(t)$ determines whether (or how much) noise to add to the training data at epoch $t$. Unlike standard simulated annealing, where $N(t)$ is always active, we choose an interleaved schedule, which injects noise for $L$ epochs after $P$ epochs without noise:
\begin{equation}
\label{eq:injection}
    N(t) = I(t) \sigma_t \ \text{with}\ I(t)= 
    \begin{cases} 
    1\ \text{if } t \ \text{mod}\ (P+L) \geq P & (\text{noise injection}) \\
    0\ \text{otherwise}  & (\text{clean data})
    \end{cases}
\end{equation}
where $\sigma_t$ represents the amount of noise, and $I(t)$ is a square-wave indicator function for the injection.

\subsubsection{Impulse noise}

Assuming normalized input data $\mathbf{x} \in [0, 1]^D$, the corrupted sample $\tilde{\mathbf{x}}_t$ is generated via a stochastic mask:
\begin{equation}
    \tilde{\mathbf{x}}_t = (1 - \mathbf{m_t}) \odot \mathbf{x} + \mathbf{m_t} \odot \mathbf{n},
\end{equation}
where $\mathbf{n} \sim \text{Bernoulli}(\tfrac{1}{2})$ represents the impulse noise $\{0,1\}$, and $\mathbf{m_t} \in \{0, 1\}^D$ is the binary corruption mask sampled element-wise such that $\mathbf{m}_t \sim \text{Bernoulli}(\sigma_t)$.
That means, $\sigma_t$ indicates the per-pixel probability of receiving a noise corruption that removes all information of the initial pixel value.

\paragraph{Expected perturbation}
The perturbation of each sample image $\mathbf{x}$ is given by;
\begin{align}
    \boldsymbol{\delta} &= \tilde{\mathbf{x}} - \mathbf{x} = \mathbf{m_t} \ \odot (\mathbf{n} - \mathbf{x}) 
\end{align}
Because the noise instance and the mask are uncorrelated, the expected perturbation over a full batch is the expected probability of pixel corruption, multiplied by the expected difference between the clean and noisy image:
\begin{equation}
    \nonumber 
    \mathbb{E}[\boldsymbol{\delta}] =  \mathbb{E}[\mathbf{m_t}] \cdot \left(\mathbb{E}[\mathbf{n} - \mathbf{x}] \right) = I(t)\sigma_t\mathbb{E}[(\tfrac{1}{2} - \mathbf{x})]
\end{equation}

To see the approximate effect the noise-injection has on the optimization trajectory, we take a second-order Taylor expansion of the loss function $\mathcal{L}(\theta, \tilde{\mathbf{x}})$ around the clean training sample $\mathbf{x}$:
\begin{equation}
    \label{eq:taylor}
    \mathcal{L}(\theta, \tilde{\mathbf{x}}) \approx \mathcal{L}(\theta, \mathbf{x}) + \boldsymbol{\delta}^\top \nabla_\mathbf{x} \mathcal{L}(\theta, \mathbf{x}) + \frac{1}{2}\boldsymbol{\delta}^\top \nabla_\mathbf{x}^2\mathcal{L}(\theta, \mathbf{x})\boldsymbol{\delta}
\end{equation}
We can take the expectation over a mini-batch,
\begin{equation}
    \label{eq:taylor}
    \mathbb{E}_{x,\boldsymbol{\delta}}[\mathcal{L}(\theta, \tilde{\mathbf{x}})] \approx \underbrace{\mathbb{E}_{\mathbf{x}}\left[\mathcal{L}(\theta, \mathbf{x})\right]}_{\text{clean loss}} + \underbrace{I(t)\sigma_t\mathbb{E}_{\mathbf{x}}\left[(\tfrac{1}{2}-\mathbf{x})^\top \nabla_x \mathcal{L}(\theta, \mathbf{x})\right]}_{\text{Jacobian regularization}} + \underbrace{\frac{I(t)\sigma_t}{2}\mathbb{E}_{\mathbf{x}}\left[\sum_i [\mathbf{H_x}]_{ii}(\tfrac{1}{2} - \mathbf{x}_i + \mathbf{x}_i^2)\right],}_{\text{weighted curvature regularization}} 
\end{equation}
which yields three separate components: the loss function of the clean data $\mathbf{x}$, a regularization term proportional to both the magnitude of the noise level $\sigma_t$ and the input sensitivity of the model to the data $\nabla_\mathbf{x} \mathcal{L}(\theta, \mathbf{x})$, and a loss term proportional to the weighted summation of the Hessian diagonal (curvature) which again is modulated by the noise level. 
%We also note that while the initial Taylor expansion assumes a small perturbation, our injected impulse noise creates pixel-by-pixel perturbation on the order of the data sample itself. But when taking the expectation over a mini-batch, the expected perturbation indeed becomes small. 
Because the mask entries are independent, the off-diagonal second-order terms scale as $\sigma_t^2$ compared to linear scaling with $\sigma_t$ of the diagonal terms, hence we drop them in \autoref{eq:taylor}.\footnote{Note the maximum $\sigma_t$ we use is 0.65, so that, for large enough batch sizes, the off-diagonal terms vanish in expectation.}

\paragraph{First order regularization} 
With Cauchy-Schwarz we can provide an upper-bound on our first-order regularization term:
\begin{equation}
    \label{eq:regulariser}
    \mathbb{E}_{\mathbf{x}}\left[(\tfrac{1}{2}-\mathbf{x}])^\top \nabla_x\mathcal{L}(\theta, \mathbf{x})\right] \leq \mathbb{E}_{\mathbf{x}}\big[\|\tfrac{1}{2} - \mathbf{x} \|_2\ \cdot \| \nabla_x \mathcal{L}(\theta, \mathbf{x})\|_2 \big]
\end{equation}
Training with impulse noise invokes a magnitude-modulated Jacobian regularization, which has been shown to improve robustness against common data corruptions and improve the identification of large scale features in deep neural networks \citep{jakubovitz2018improving,chan2019jacobian, hoffman2019robust, cohen2019certified}. The magnitude of the first-order regularizer is maximized for high contrast $x_i \in\{0,1\}$ pixels and zero for gray $x_i=0.5$ pixels. This means our regularizer will more aggressively reduce the sensitivity to high contrast (extreme pixel values) features, while retaining high sensitivity for low contrast features.

\paragraph{Second order regularization}
We can also upper-bound the second-order regularization term, connecting the minimization of the trace to minimizing the spectral norm (sharpness) of the Hessian matrix:
\begin{equation}
    \boldsymbol{\delta}^\top \mathbf{H_x}\boldsymbol{\delta} \leq \lambda_{\rm{max}}(\mathbf{H_x})\|\boldsymbol{\delta} \|^2_2
\end{equation}

\begin{equation}
    \mathbb{E}_{\boldsymbol{\delta}|\mathbf{x}}\left[\boldsymbol{\delta}^\top \mathbf{H_x}\boldsymbol{\delta} \right] \leq \lambda_{\rm{max}} \cdot\sigma_t\sum_i(\tfrac{1}{2} - \mathbf{x}_i + \mathbf{x}_i^2)
\end{equation}
By minimizing the expected local sharpness in \autoref{eq:taylor}, impulse noise injection implicitly minimizes the sharpness (spectral norm) weighted by pixel-level contrast, ideally leading to flatter, more generalizable solutions \citep{hochreiter1997flat}. This is similar to regularization approaches suggested in Sharpness-Aware-Minimization \citep{foret2020sharpness}, but it is important to point out that we minimize the sharpness w.r.t. the input data $\mathbf{x}$ instead of the weights $\boldsymbol\theta$. Because impulse noise stochastically removes local information, the training is thus regularized against overfitting to high-frequency spatial features. Instead, the training encourages learning large-scale features that remain robust even to high levels of impulse noise. 

%\paragraph{Caveats} While this derivation is useful to get an idea of how our salt and pepper noise affects the optimization routine, we must point out that the perturbation $\boldsymbol{\delta}$ may be quite large, in fact it may approach the magnitude of the data itself. 

\subsubsection{Gaussian noise schedule}
For interleaved Gaussian noise, we sample from a zero-mean Gaussian distribution, $\tilde{\mathbf{x}}_t = \mathbf{x} + \mathbf{n}_t$, where $\mathbf{n}_t\sim\mathcal{N}(\mathbf{0}_D, \sigma_t\mathbf{1}_D)$. The expected perturbation is thus $\mathbb{E}[\boldsymbol{\delta}] = 0$, and the expectation of the loss function over a mini-batch reduces to
\begin{equation}
    \label{eq:taylor-gauss}
    \mathbb{E}_{x,\boldsymbol{\delta}}[\mathcal{L}(\theta, \tilde{\mathbf{x}})] \approx \underbrace{\mathbb{E}_{\mathbf{x}}\left[\mathcal{L}(\theta, \mathbf{x})\right]}_{\text{clean loss}} + \underbrace{\frac{I(t)\sigma_t^2}{2} \mathbb{E}_{\mathbf{x}}\left[\rm{\mathbf{Tr}}(\mathbf{H_x})\right]}_{\text{curvature regularization}}.
\end{equation}
We no longer get a Jacobian regularization term but retain the familiar second-order curvature regularization, whose strength is again modulated by the noise curriculum. This term results in a uniform curvature penalty, independent of local pixel contrast.

\subsection{Gradient-norm stabilization}
\label{sec:gradnorm}
The general update rule (ignoring preconditioning terms) for SGD-type algorithms is: 
\begin{align}
   \mathbf{g}_t = \eta_t \nabla_\theta \mathcal{L}(\theta_{t}, \mathbf{x})
\end{align}
where $\eta_t$ is the time-dependent learning rate. A potential issue with our interleaved noise schedule is that for a constant learning rate, the magnitude of the noisy gradient update $\tilde{\mathbf{g}}_t$ may deviate significantly from $\mathbf{g}_t$.

For example, if we were to introduce an image of pure noise, the magnitude of the gradient term $\nabla_\theta \mathcal{L}(\theta_t, \mathbf{\tilde{x}})$ would represent an average of $n$ random vectors where $n$ is the batch size. As $n \to \infty$ this would cause $\nabla_\theta \mathcal{L} \to 0$. Alternatively, at late training times when $\nabla_\theta \mathcal{L}(\theta_t, \mathbf{x})$ is small due to descending into a minima, even a small perturbation to $\mathbf{x}$ due to the impulse noise could lead to significant increases in the gradient norm. Hence, we modulate the noisy update magnitude $\tilde{\mathbf{g}}_t$ with a scalar prefactor term $p_t$, defined as
\begin{align}
\label{eq:stabalisation}
   p_t = f \cdot \frac{\|\nabla_\theta \mathcal{L}(\theta_{t-1}, \mathbf{x}) \|_2 }{\|\nabla_\theta \mathcal{L}(\theta_t, \tilde{\mathbf{x}}) \|_2},
\end{align}
where the magnitude of the gradient norm of the last clean update (averaged over a full epoch) determines the magnitude of the current noisy update. We also introduce a tuning parameter $f$, which scales the overall strength of the noisy gradient updates (\autoref{sec:grad-scaling} investigates the effect of this parameter). The update step during the noisy epochs is then
\begin{align}
   \tilde{\mathbf{g}}_t = \eta_t p_t\nabla_\theta \mathcal{L}(\theta_{t}, \tilde{\mathbf{x}}).
\end{align}

\section{Experiments}\label{sec:experiments}
We evaluate the proposed interleaved curriculum using both impulse and Gaussian noise 1) against the standard training on clean data, 2) in comparison to alternative noise injection curricula, and 3) against more complex noise injection and augmentation methods, specifically:
\begin{enumerate}
    \item \textbf{Cutout} \citep{devries2017improved} to determine whether generalization benefits from noise injection exceed those from masking large patches;
    \item \textbf{Sharpness-Aware Minimization} \citep[SAM,][]{foret2020sharpness} to test implicit input-space curvature minimization against explicit weight-space curvature minimization;
    \item \textbf{AugMix} \citep{hendrycks2019augmix}, which performs a series of structural augmentations to the input images, to determine benefits for corrupted and OOD data errors;
    \item \textbf{Vital Phase Augmentation} \citep[VIPAug,][]{lee2024domain}, a novel approach that introduces phase shifts during training, to compare frequency-space augmentation to spatial-domain regularization.
\end{enumerate}
In each case, we measure the performance on clean, corrupted, and OOD datasets. 

\paragraph{Architectures and training details}
For our CIFAR-100 experiments, we utilize a WideResNet architecture \citep{zagoruyko2016wide}. For ImageNet-1k evaluations, we test both ResNet50 \citep{he2016deep} and Vision Transformer \citep[ViT,][]{dosovitskiy2020image} architectures to ensure our findings generalize across convolutional and attention-based models. We train the WideResNet from scratch for 100 epochs; for the ResNet50 and ViT we begin with pre-trained models and perform fine-tuning for 50 and 30 epochs for the ResNet50 and ViT, respectively. The ResNet50 architecture follows from \citet{he2016deep} and we use the ViT-B/16 variant with a patch size of $16 \times 16$ and an embedding dimension of 768.
We perform hyper-parameter tuning over the learning rate for all results reported.  For the CIFAR-100 trials our baseline method is AdamW, while for ImageNet-1k we use SGD with momentum for the ResNet50 model and AdamW for the ViT. For AugMix we use identical hyperparameters for the severity and Jensen-Shannon divergence loss as reported in \citep{hendrycks2019augmix}.

For our noise schedules, we inject impulse or Gaussian noise to all training samples for one full epoch, once every 5 epochs, in all trials $(P=5)$\footnote{Barring the results shown in Figures 1, and 2 which use $P=10$ and $L=5$, but are for visualization purposes and not included in any tabulated results.}. We note that this results in a square wave curriculum with $156$ noisy gradient updates every 5 epochs for the CIFAR-100 trials using a batch size of 256, and $8008$ noisy updates for both ImageNet trials which use a batch size of 128. For all reported results we use the gradient norm scaling factor $f=0.4$, which was approximately the best performing trial in each case. We explicitly avoid a large hyperparameter sweep over $P$ and $f$ for our presented results to show the practical usefulness of this method without the need for extra costly tuning. Preliminary ablation tests (see Sections \ref{sec:ablations}) showed performance was robust for a wide range of $P$ and $f$.

We train all models on NVIDIA-A100 GPUs. The training times remain unchanged because our method (unlike SAM, AugMix or VIPAug) does not require any additional forward or backwards passes over the data nor any additional clean epochs. The only extra computational cost stems from tracking the last clean gradient norm and calculating the gradient-norm scaling in \autoref{eq:stabalisation}.

\paragraph{Corruption and robustness testing}
To assess model robustness against common corruptions, we use the CIFAR-100-C and ImageNet-C benchmark datasets introduced by
\citet{hendrycks2019benchmarking}. These datasets apply 15 distinct types of corruptions across five severity levels.
To test our noise injection method for improvements in OOD performance, we compare the inference performance of the ResNet50 and ViT models, trained on the ImageNet-1k dataset, on the ImageNet-R dataset \citep{hendrycks2021many}. This dataset consists of 200 subclasses of ImageNet-1k which have been altered to test for out-of-distribution robustness.
We report standard clean top-1 error, mean corruption error (mCE), and structural mCE (which removes the noise-based corruption classes). All results are averaged over 5 random seeds with bootstrapped errors shown.

\section{Results}

We report error rates ($\downarrow$ is better) in all tables and accuracy ($\uparrow$ is better) in all figures.

\subsection{Comparison to clean training and other noise injection curricula}

For this section, we train the WideResNet model on CIFAR-100 data, using either only clean, or only noisy, or a mix of clean and noisy data, time-ordered as an interleaved curriculum ($P=10, L=5$; see \autoref{eq:injection}). The dataset was split into a 70/10/20 train/validation/test split, and each dataset was split further into a clean and noisy set, 
%with 80$\%$ of the data used for the clean split
where the latter had impulse noise injected at a level of $\sigma=0.65$. 

\autoref{fig:clean_vs_injected} shows the accuracy on clean and noisy test data. It is evident that dataset switches are quickly picked up by the model: the transition from clean to noisy initially sharply degrades clean test accuracy, but only for a short period, while producing significant benefits in the long run. In particular: \textbf{Training with interleaved noise improves long-term clean test accuracy beyond training on clean data only}, even though the model probes the clean data for fewer epochs. We interpret this outcome as the result of increased robustness during training with noisy data, leading to minimizers with wider, shallower minima \citep{hendrycks2021many,spiking_noise,ly2025Nature}. 
On the flip side, the interleaved scheme also shows drastic improvement on noisy data: When the training is stopped during a noise-injection period, \textbf{the long-term accuracy on noisy data outperforms training on noisy data only.} We interpret this behavior as the benefit of having access to high-quality samples even if the target data distribution is noisy.

\begin{figure*}[t]
    \centering
    \includegraphics[width=0.95\linewidth]{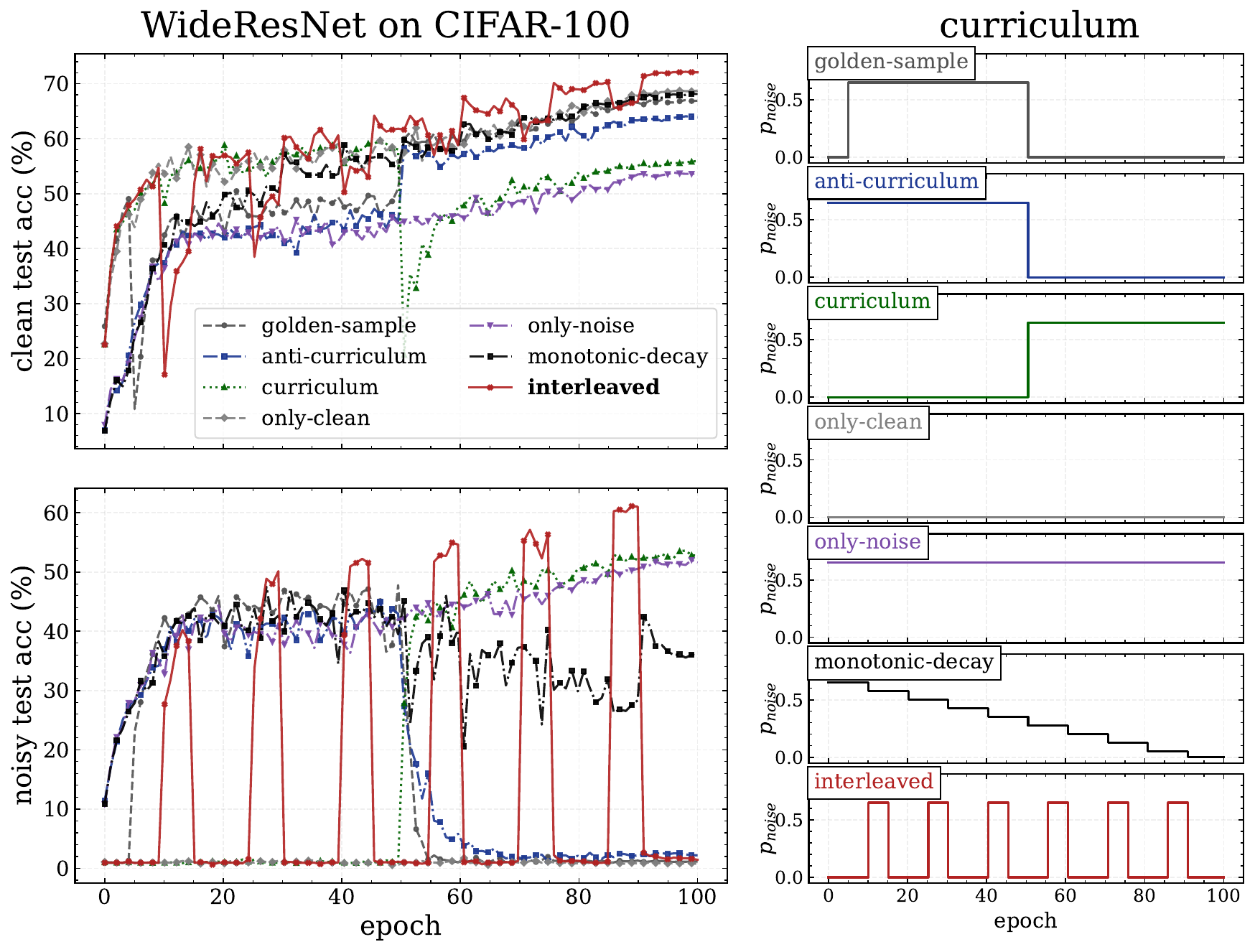}
    \caption{Comparison of clean (top) and noisy (bottom) test accuracy throughout training for seven noise injection curricula (right panels): golden-sample, curriculum, anti-curriculum, only-clean, only-noise, monotonic-decay, and an example of an interleaved schedule with $P=10, L=5$.}
    \label{fig:curriculum}
\end{figure*}

The last result has implications for curriculum and transfer learning, which often employ only one or a small number of sharp transitions in dataset quality or optimization objective.
We therefore expand the above test to include a suite of proposed curricula for noise injection. 
\autoref{fig:curriculum} shows that the interleaved schedule consistently achieves the highest final accuracy on both the clean and noisy test datasets. Common alternative curricula either lose their regularization capacity late in training, or fail to converge on the clean data in the finite amount of epochs. We interpret these results as the benefits of repeated ``kicks'' out from narrow minima into wider ones as well as the repeated ``reminders'' of what accurate data looks like.

\subsection{Comparison to other augmentation methods}

The previous section demonstrated the effectiveness of interleaved noise for a specific model and interleaved period ($P=10, L=5$), which may appear fine-tuned. In this section we demonstrate that a fixed, simple choice of a single noisy epoch for every five clean epochs, i.e. $P=5, L=1$, reaps the same benefits, even when compared to more advanced and computationally costly augmentation schemes. An ablation study on the best choice for $P$ is presented in \autoref{sec:ablation-p}.

\subsubsection{CIFAR-100}
\label{sec:cifar100}

%--------------------
% cifar100 full table
\begin{table}[t]
\caption{\textbf{CIFAR-100 Results.} Comparison of robustness methods on WideResNet trained on CIFAR-100 and evaluated on CIFAR-100-C. We report Clean Error, Mean Corruption Error (mCE), and structural mCE (excluding noise). \textbf{Bold} indicates the best result; \underline{underline} indicates the second best. Improvements over the respective baseline are shown in \textcolor{darkgreen}{(green)}, $\pm$ values denote bootstrap standard errors. }
\label{tab:cifar_results}
\begin{center}
\begin{small}
\setlength{\tabcolsep}{10pt}
\begin{tabular}{l lll}
\toprule
Method & Clean Err ($\downarrow$) & mCE ($\downarrow$) & Struct. mCE  ($\downarrow$) \\
\midrule 
\textcolor{black!70}{AdamW} & \textcolor{black!70}{25.7 $\pm$ 0.3} & \textcolor{black!70}{52.6 $\pm$ 0.4} & \textcolor{black!70}{47.1 $\pm$ 0.4} \\
AdamW + Gaussian & 24.9 $\pm$ 0.3 \improv{-0.8} & 47.5 $\pm$ 0.4 \improv{-5.1} & 43.2 $\pm$ 0.3 \improv{-3.9} \\
 AdamW + Impulse& 24.8 $\pm$ 0.4 \improv{-0.9} & 49.3 $\pm$ 0.3 \improv{-3.3} & 43.8 $\pm$ 0.3 \improv{-3.3} \\
\midrule
\textcolor{black!70}{Cutout \citep{devries2017improved}} & \textcolor{black!70}{24.9 $\pm$ 0.3} & \textcolor{black!70}{ 52.6 $\pm$ 0.3} & \textcolor{black!70}{46.4 $\pm$ 0.2} \\
Cutout + Gaussian & 24.9 $\pm$ 0.2 \improv{0.0} & 47.9 $\pm$ 0.3 \improv{-4.7} & 43.4 $\pm$ 0.3 \improv{-3.0} \\
Cutout + Impulse & \textbf{24.3 $\pm$ 0.1} \improv{-0.5} & 49.1 $\pm$ 0.4 \improv{-3.5} & 42.5 $\pm$ 0.4 \improv{-3.9} \\
\midrule
\textcolor{black!70}{SAM \citep{foret2020sharpness}} & \textcolor{black!70}{25.0 $\pm$ 0.4} & \textcolor{black!70}{52.3 $\pm$ 0.2} & \textcolor{black!70}{46.2 $\pm$ 0.3} \\
SAM + Gaussian & 24.8 $\pm$ 0.3 \improv{-0.2} & 46.3 $\pm$ 0.2 \improv{-6.0} & 43.7 $\pm$ 0.3 \improv{-2.5} \\
SAM + Impulse & 24.7 $\pm$ 0.3 \improv{-0.3} & 48.2 $\pm$ 0.3 \improv{-4.1} & 42.4 $\pm$ 0.3 \improv{-3.8} \\
\midrule
\textcolor{black!70}{VIPAug \citep{lee2024domain}} & \textcolor{black!70}{25.7 $\pm$ 0.4} & \textcolor{black!70}{52.4 $\pm$ 0.2} & \textcolor{black!70}{46.8 $\pm$ 0.3} \\
VIPAug + Gaussian & 25.2 $\pm$ 0.2 \improv{-0.5} & 46.5 $\pm$ 0.2 \improv{-5.9} & 42.8 $\pm$ 0.2 \improv{-4.0} \\
VIPAug + Impulse & 24.9 $\pm$ 0.3 \improv{-0.8} & 49.6 $\pm$ 0.3 \improv{-2.8} & 43.0 $\pm$ 0.3 \improv{-3.8} \\
\midrule
\textcolor{black!70}{AugMix \citep{hendrycks2019augmix}} & \textcolor{black!70}{25.2 $\pm$ 0.2} & \textcolor{black!70}{36.9 $\pm$ 0.2} & \underline{\textcolor{black!70}{34.6 $\pm$ 0.2}} \\
AugMix + Gaussian & 24.9 $\pm$ 0.2 \improv{-0.3} & \textbf{35.3 $\pm$ 0.4} \improv{-1.6} & \textbf{33.9 $\pm$ 0.4} \improv{-0.7} \\
\rowcolor{green!10} AugMix + Impulse & \textbf{24.3 $\pm$ 0.4} \improv{-0.9} & \underline{36.5 $\pm$ 0.1} \improv{-0.4} & \textbf{33.9 $\pm$ 0.1} \improv{-0.7} \\
\bottomrule
\end{tabular}
\end{small}
\end{center}
\end{table}

%-----------------------
% minimal table cifar100
%\input{cifar_minimal_table}

% cifar severity figure
%\begin{figure}
%    \centering
%    \includegraphics[width=0.92\linewidth]{figures/money_category_severity.pdf}
%    \caption{CIFAR100-C accuracy across increasing corruption severalties (1–5) for specific corruption categories (Noise, Blur, Weather, Digital), we show the CIFAR100 (clean) performance at a severity level 0. Our interleaved SP curriculum (purple) demonstrates superior robustness preservation especially for the weather corruptions.}
%    \label{fig:cifar-severity}
%\end{figure}

We compare the performance of our method on the WideResNet model and the CIFAR-100 and CIFAR-100-C datasets. The baseline is a standard AdamW training run, which is enhanced with either Gaussian or impulse noise injection. The alternative methods use training with Cutout, SAM, VIPAug, and AugMix augmentations.
\autoref{tab:cifar_results} shows that the inclusion of interleaved Gaussian or impulse noise improves clean accuracy,  mCE, and structural mCE for all methods, when the right noise distribution is selected. The best performance comes from using AugMix with interleaved impulse noise, which reduced the clean error by $1.4\%$, mCE by $16.1\%$, and structural mCE by $13.2\%$ compared to the baseline AdamW run. We note that much of this improvement for the corruption robustness comes from AugMix alone, but the noise injection further improves the results by $0.9\%$, $0.4\%$ and $0.7\%$ across clean, mCE and structural mCE, respectively. But even just our simple noise  injection method improves performance to levels comparable or better than these more expensive schemes.

%We also note, that in all barring the vanilla AdamW trials, the Gaussian noise reduces the mCE more than using S$\&$P noise, however this relative reduction in corruption accuracy is heavily dominated by test images in the ``noise'' category of ImageNet-C. When we remove the ``noise'' category, which we do in the Struct. mCE metric, we see the S$\&$P discrete tempering trials match or outperform the Gaussian trials. For all baseline methods, S$\&$P noise curricula reduce the clean error the most.

%We break down both the corruption type, and corruption severity level in \autoref{fig:cifar-severity}, and show the performance of all baselines methods, as well as AdamW and Augmix with the interleaved SP noise. Perhaps surprisingly, we see discrete tempering shows the best improvement for the weather-type corruptions \citep[see][for details and visualizations of corruption types]{hendrycks2019benchmarking}, with little improvement for the averaged noise corruptions with plain AugMix performing better for high severity noise.

\subsubsection{ImageNet-1k}
%--------------------
% imagenet main table
\begin{table}[t]
\caption{\textbf{ImageNet Results}. \textbf{Struct}: Structural mCE (excluding noise category). \textbf{Rend}: ImageNet-R Rendition Error (OOD). All metrics are error percentages ($\downarrow$), with bootstrapped standard errors.}
\label{tab:imagenet_results}
\begin{center}
\small
\begin{tabular}{ll llll}
\toprule
Architecture & Method & Clean Err ($\downarrow$) & mCE ($\downarrow$) & Struct. ($\downarrow$) & Rend. ($\downarrow$)\\
\midrule
\multirow{16}{*}{ResNet50} 
& SGD              & $25.8 \pm.1$ \phantom{\scriptsize{$-0.0$}} & $89.1\pm.2$ \phantom{\scriptsize{$-0.0$}} & $88.1\pm.2$ \phantom{\scriptsize{$-0.0$}} & $63.2 \pm.2$ \phantom{\scriptsize{$-0.0$}} \\
& SGD + Gaus.   & $25.5 \pm$.2 \improv{-0.3} & $88.2\pm$.2 \improv{-0.9} & $87.2\pm$.1 \improv{-0.9} & $62.4\pm$.1 \improv{-0.8} \\
& SGD + Impulse         & $25.1\pm.2$ \improv{-0.7} & $87.3\pm$.1 \improv{-1.8} & $86.2\pm$.1 \improv{-1.9} & $61.8\pm$.1\improv{-1.4} \\ \cmidrule{2-6}
& Cutout              & 24.4$\pm$.1 \phantom{\scriptsize{$-0.0$}} & 88.2$\pm$.3 \phantom{\scriptsize{$-0.0$}} & 87.0$\pm$.3 \phantom{\scriptsize{$-0.0$}} & 61.6$\pm$.1 \phantom{\scriptsize{$-0.0$}} \\
& Cutout + Gaus.   & 24.5$\pm$.1 \worse{+0.1} & 88.0$\pm$.2 \improv{-0.2} & 86.9$\pm$.2 \improv{-0.1} & 62.5$\pm$.4 \worse{+0.9} \\
& Cutout + Impulse         & 23.7$\pm$.1 \improv{-0.7} & 86.5$\pm$.3 \improv{-1.7} & 85.4$\pm$.4 \improv{-1.6} & 60.8$\pm$.4 \improv{-0.8} \\ 
\cmidrule{2-6}
& SAM               & $24.9 \pm.2$ & $88.3 \pm.2$ & $87.5 \pm.2$ & $62.6 \pm.1$ \\
& SAM + Gaus.    & $24.7 \pm.1$ \improv{-0.2} & $85.7 \pm.1$ \improv{-2.6}& $86.4 \pm.1$ \improv{-1.1}& $61.8 \pm.2 $ \improv{-0.8}\\
& SAM + Impulse          & $24.8 \pm.1$ \improv{-0.1}& $84.8 \pm.1$ \improv{-3.5}& $84.6 \pm.3$ \improv{-2.9}& $60.5 \pm.3$ \improv{-2.1}\\ 
\cmidrule{2-6}
& VIPAug           & $28.7\pm.1$ & $89.7 \pm.2$ & $88.6 \pm.2$ & $66.0 \pm.1$ \\
& VIPAug + Gaus.    & $ 28.6\pm.3$ \improv{-0.1}& $89.6\pm.3$ \improv{-0.1}& $88.5 \pm.3$ \improv{-0.1}& $65.9\pm.1$ \improv{-0.1}\\
& VIPAug + Impulse       & $28.1 \pm.3$ \improv{-0.6}& $88.6 \pm.2$ \improv{-1.1}& $87.5 \pm.1$ \improv{-1.1}& $64.7 \pm.1$ \improv{-1.3}\\
\cmidrule{2-6}
& AugMix            & $23.8 \pm.2$ & $83.2 \pm.2$ & $81.6 \pm.2$ & $55.3 \pm.1$ \\
& AugMix + Gaus.    & $23.7 \pm.2$ \improv{-0.1} & $82.3\pm.1$ \improv{-0.9} & $80.7\pm.1$ \improv{-0.9} & $54.6\pm.1$ \improv{-0.7} \\
& AugMix + Impulse       & $\mathbf{23.3\pm.2}$ \improv{-0.5}& $\mathbf{82.0\pm.1}$ \improv{-1.2} & $\mathbf{80.3\pm.1}$ \improv{-1.3} & $\mathbf{54.7\pm.1}$ \improv{-0.6} \\
\midrule \midrule
\multirow{16}{*}{ViT}
& AdamW              & 20.1$\pm$.3 \phantom{\scriptsize{$-0.0$}} & 69.5$\pm.4$ \phantom{\scriptsize{$-0.0$}} & $67.2\pm.3$ \phantom{\scriptsize{$-0.0$}} & $55.9\pm.3$ \phantom{\scriptsize{$-0.0$}} \\
& AdamW + Gaus.   & 19.5$\pm.2$ \improv{-0.6} & 66.1$\pm.1$ \improv{-3.4} & $64.3\pm.2$ \improv{-2.9} & $53.8\pm.2$ \improv{-2.1} \\
& AdamW + Impulse         & $20.0\pm.2$ \improv{-0.1} & $67.7\pm.3$ \improv{-1.8} & $65.7\pm.4$ \improv{-1.5} & $54.7\pm.3$ \improv{-1.2} \\ \cmidrule{2-6}
& Cutout              & $19.9 \pm.1$ & $69.4 \pm.4$ & $67.2 \pm.4$ & $55.7 \pm.1 $ \\
& Cutout + Gaus.   & $19.2\pm.1$ \improv{-0.7}& $66.3\pm.2$ \improv{-3.1}& $64.5\pm.1$ \improv{-2.7}& $53.4\pm.1$ \improv{-2.3}\\
& Cutout + Impulse         & $20.3\pm.1$ \worse{+0.4}& $68.8\pm.4$ \improv{-0.6}& $66.7\pm.1$ \improv{-0.5}& $55.7\pm.1$ \improv{-0.0}\\ 
\cmidrule{2-6}
& SAM               & $17.5 \pm.1 $ & 63.9 $\pm$.2 & $62.4 \pm.1$ & $50.5 \pm.1$ \\
& SAM + Gaus.    & $\mathbf{17.4 \pm.1}$ \improv{-0.1} & $63.1 \pm.1$ \improv{-0.8} & $62.1\pm.1$ \improv{-0.3} & $50.2 \pm.1$ \improv{-0.3}\\
& SAM + Impulse          & $17.7 \pm.1$ \worse{+0.2} & $63.6 \pm.1$ \improv{-0.3}& $62.7 \pm.1$ \worse{+0.3}& $50.6 \pm.1$ \worse{+0.1} \\ 
\cmidrule{2-6}
& VIPAug            & $19.8 \pm.2$ & $69.8 \pm.6$ & $67.6 \pm.2$ & $55.2 \pm.1$ \\
& VIPAug + Gaus.    & $18.9 \pm.1$ \improv{-0.9}& $ 66.2\pm.2$\improv{-3.6} & $ 64.4\pm.1$ \improv{-3.2}& $ 53.5\pm.1$ \improv{-1.7} \\
& VIPAug + Impulse       & $20.0 \pm.1$ \worse{+0.2}& $68.7 \pm.5$\improv{-1.1} & $ 66.8\pm.3$ \improv{-0.8}& $54.1 \pm.4$ \improv{-1.1}\\
\cmidrule{2-6}
& AugMix            & $17.8 \pm.1$ & $62.4 \pm.1$ & $60.7 \pm.1$ & $48.6 \pm.1$ \\
& AugMix + Gaus.    & $17.5 \pm.1$ \improv{-0.3}& $ \mathbf{60.8\pm.1}$ \improv{-1.6}& $\mathbf{59.7 \pm.1}$ \improv{-1.0}& $ 48.6\pm.1$ \improv{-0.0}\\
& AugMix + Impulse       & $17.5 \pm.1$ \improv{-0.3}& $ 61.6\pm.1$ \improv{-0.8}& $60.1 \pm.1$ \improv{-0.6}& $ \mathbf{48.3\pm.1}$ \improv{-0.3}\\
\bottomrule
\end{tabular}
\end{center}
\end{table}

We now investigate if these results hold up for larger models by fine-tuning pretrained ResNet50 and ViT architectures on the ImageNet-1k dataset. We report the clean error, corruption, and structural corruption error as well as the OOD accuracy via testing on the ImageNet-R dataset \citep{hendrycks2021many}. 
\autoref{tab:imagenet_results} shows that for the overwhelming majority of baseline regularization methods, the addition of either Gaussian or impulse noise with our interleaved curriculum reduces the error across all four performance metrics. Interestingly, for the ResNet50 models the impulse noise appears to work best to reduce the clean, corrupt, structural, and OOD error rates, while for the ViT Gaussian noise performs significantly better across all baselines. We will explore the  reasons for this architecture dependence in \autoref{sec:architecture}.

% -------------------------
% imagenet corruption types
\begin{figure}
    \centering
    \includegraphics[width=0.95\linewidth]{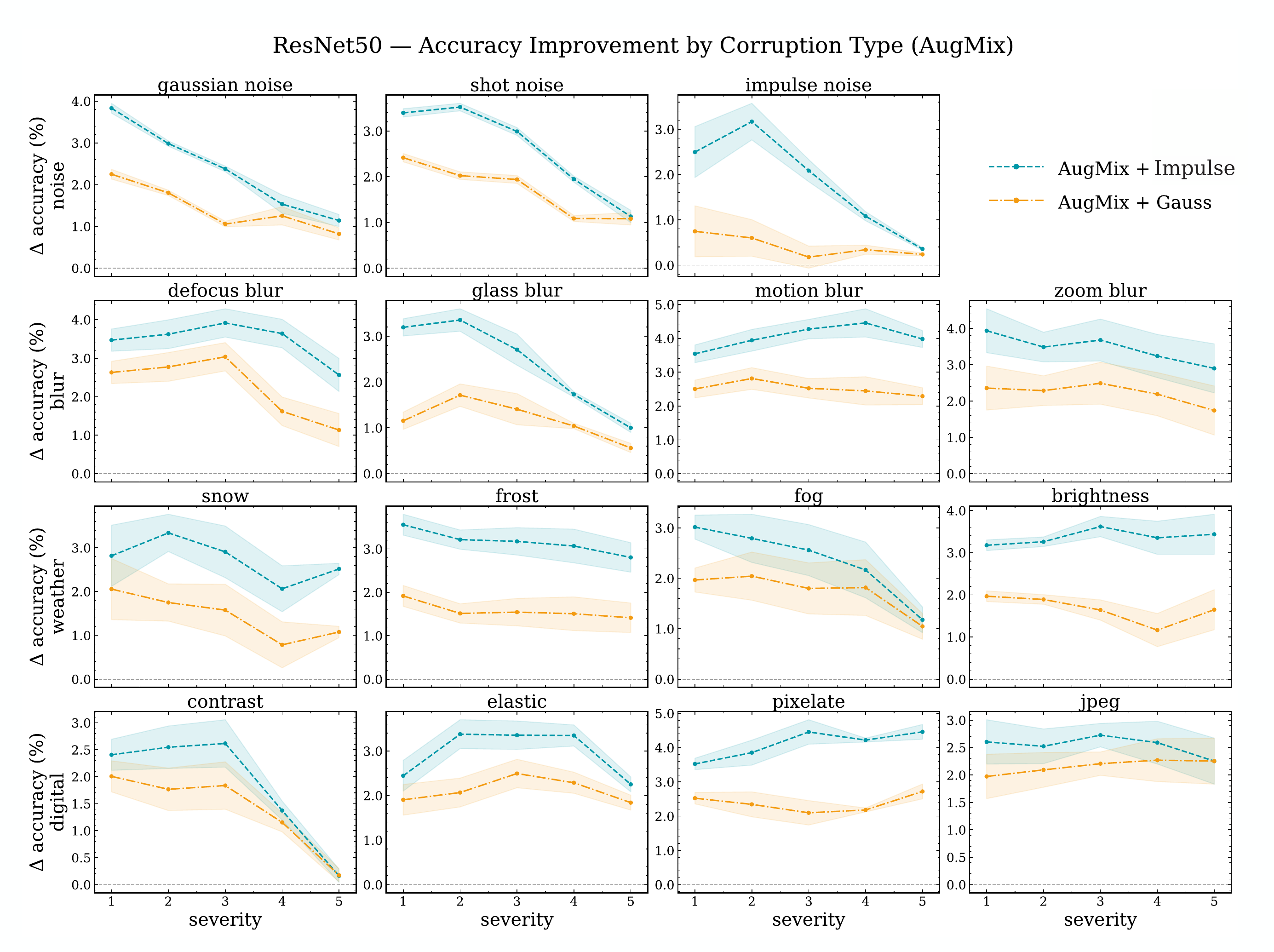}
    \caption{\textbf{Results for different corruptions.}  Accuracy improvement from noise injection with impulse (blue) and Gaussian noise (orange) over the reference AugMix for the ResNet50 model on the ImageNet-C dataset \citep{hendrycks2019benchmarking}, sorted by corruption category (rows) and subcategory (columns).}
    \label{fig:imagenet-corruption}
\end{figure}

In \autoref{fig:imagenet-corruption} we show a breakdown of the test performance by the specific type of corruption from \citet{hendrycks2019benchmarking}. For a clearer comparison, we only show the best performing baseline regularizer with and without noise injection (full results of per-type corruption accuracy are listed in \autoref{sec:performance-severity}). Injection of impulse noise outperforms both baseline AugMix and the Gaussian noise injection at every severity level across all metrics.

%We explore the difference between the type of noise used in the discrete tempering further in \autoref{sec:architecture}. We note the relatively poor clean performance from using VIPAug with the ResNet50 model. This is likely due to the training time of the fine-tuning being insufficient to adapt the running BatchNorm statistics to the shifted activation distribution caused but VIPAug's phase perturbation. However, the main focus here is the improvement made to the baseline method when applying discrete tempering which is well demonstrated. We do not see the same performance degradation of VIPAug in the end-to-end trained Cifar100 trials, or the pretrained ViT (which does not perform batch normalization).

\begin{figure}
    \centering
    \includegraphics[width=0.99\linewidth]{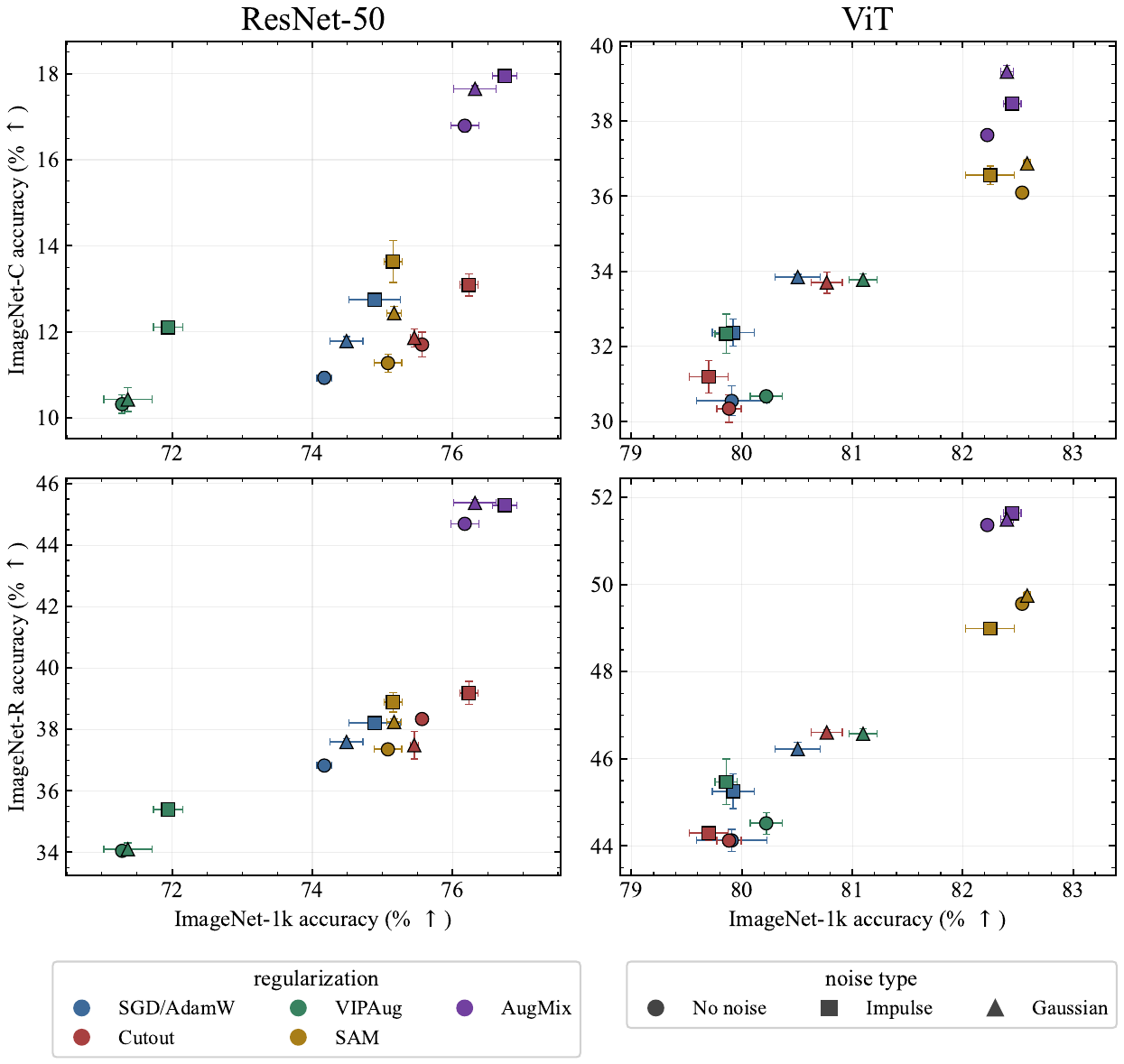}
    \caption{Clean accuracy vs. corruption and OOD robustness trade-off for ResNet50 and ViT on ImageNet-1k. Each point represents a distinct trial, with clean accuracy on the x-axis and mean corruption accuracy (top row) and ImageNet-R OOD accuracy (bottom row) on the y-axis for ResNet50 (left column) and ViT (right column). Point color indicates the baseline regularization method with the shape indicating noise type. All results are averaged over 5 seeds with the error bars showing the bootstrapped standard error.}
    \label{fig:imagenet-scatterplot}
\end{figure}

We visualize the test accuracy of the same runs used for \autoref{tab:imagenet_results} in \autoref{fig:imagenet-scatterplot}, showing the performance of the ResNet50 and ViT models on the clean (ImageNet-1k), corrupt (ImageNet-C), and OOD (ImageNet-R) test sets. It confirms the clear trend of interleaved noise injection consistently improving across performance metrics and optimization baselines. %Interestingly we see for the ViT, using Gaussian noise with discrete tempering results in better results for both clean and corrupt datasets, we explore this trend more in \autoref{sec:architecture}. Overall we see in almost all trials discrete tempering regardless of noise distribution again improves results.

% imagenet panel plot to compare noise types

\subsection{Ablations}
\label{sec:ablations}
We now present ablation tests over a set of hyperparameters we have introduced in this paper.

\subsubsection{Gradient-norm scaling}
\label{sec:grad-scaling}
We perform ablations over the gradient-norm scaling factor $f$ with the WideResNet model on the CIFAR-100 and CIFAR-100-C datasets. \autoref{tab:ablation_f} shows results with varying levels of $f$ as well as a trial with no gradient-norm stabilization. We find that reducing the value of $f$ to around 0.2 tends to increase the error on the clean data, however increasing $f$ increases the performance on the corrupt data. The general trend shows that increasing $f$ past $0.4$ reduces the corruption error rates, but at the cost of higher clean error rates. For impulse noise, the clean error rate never rises above $25.4 \pm.3$. The performance of both the Gaussian and impulse noise injection with no gradient-norm stabilization degrades noticeably on clean data, however the corruption performance improves. %We reiterate for all results reported in this study, we selected hyperparameter configurations based on their performance on the clean datasets.

%\autoref{fig:grad-norm-fac} shows the clean test accuracy vs mCA for the WideResNet trained on Cifar-100. The marker styles show the type of noise and regularization, with the color indicating $f$. We show results for the AdamW baseline, as well as the best performing regularizer AugMix, and here extend the ablation of $f$ up to 2. We see a clear pareto-front forming between increasing the corruption performance (x-axis) at the cost of reducing the clean performance, a trend especially clear in the Gaussian noise trials. This reiterates the effect of $f$ is to balance the regularization strength, with the clean performance.

%We also show the effect of $f$ for different levels of noise $\sigma_t$ in \autoref{fig:noise-ablation}, again for the WideResNet on Cifar100, finding no significant differences with the effect of $f$ for different noise levels, in general clean performance degrades as $f$ increases up to 1.

\begin{table}
\centering
\caption{Ablation over the gradient-norm scaling factor $f$. We report Clean Error, Mean Corruption Error (mCE) for both impulse and Gaussian interleaved noise injection for the WideResNet model. All metrics are error percentages ($\downarrow$).}
\label{tab:ablation_f}
\begin{tabular}{c ccc ccc}
\toprule
& \multicolumn{3}{c}{\textbf{Gaussian noise}} & \multicolumn{3}{c}{\textbf{Impulse noise}} \\
\cmidrule(lr){2-4} \cmidrule(lr){5-7}
\textbf{scaling factor ($f$)} & Clean & mCE & Struct. & Clean & mCE & Struct. \\
\midrule
0.2 & $25.0 \pm.2$ & $47.2 \pm.1$ & $43.7 \pm.2$ & $25.0 \pm.2$ & $50.6 \pm.2$ & $44.5 \pm.1$ \\
0.4 & $\mathbf{24.9 \pm.1}$ & $47.5 \pm.2$ & $43.2 \pm.1$ & $\mathbf{24.9 \pm.3}$ & $49.3 \pm.1$ & $43.8 \pm.1$ \\
0.6 & $25.2 \pm.2$ & $46.1 \pm.1$ & $42.7 \pm.2$ & $25.0 \pm.1$ &$49.3 \pm.1$ & $43.0 \pm.1$ \\
0.8 & $25.4 \pm.1$ & $44.9 \pm.2$ & $42.2 \pm.2$ & $25.1 \pm.1$ & $48.9 \pm.1$ & $42.4 \pm.2$ \\
1.0 & $25.6 \pm.1$ & $45.1 \pm.1$ & $42.0 \pm.1$ & $25.0 \pm.1$ & $48.4 \pm.1$ & $42.4 \pm.1$ \\
1.2 & $26.1 \pm.1$ & $43.9 \pm.2$ & $41.9 \pm.1$ & $25.2 \pm.1$ & $48.8 \pm.2$ & $42.3 \pm.3$ \\
1.4 & $25.9 \pm.1$ & $43.7 \pm.1$ & $41.9 \pm.1$ & $25.4 \pm.2$ & $48.7 \pm.1$ & $42.4 \pm.1$ \\
1.6 & $26.4 \pm.1$ & $43.6 \pm.1$ & $\mathbf{41.8 \pm.2}$ & $25.2 \pm.1$ &$48.1 \pm.5$ & $41.8 \pm.4$ \\
1.8 & $26.6 \pm.2$ & $43.6 \pm.2$ & $42.0 \pm.1$ & $25.3 \pm.3$ & $47.9 \pm.3$ & $41.6 \pm.1$ \\
2.0 & $26.9 \pm.2$ & $\mathbf{43.5 \pm.1}$ & $41.9 \pm.1$ & $25.3 \pm.1$ & $\mathbf{47.7 \pm.2}$ & $\mathbf{41.6 \pm.2}$ \\
\midrule
\rowcolor{red!8} no stabilization & $27.6 \pm .1$ & $43.1 \pm .1$ & $41.8 \pm .1$ & $26.0 \pm .2$ & $47.2 \pm.1$ & $41.4 \pm .1$ \\
\bottomrule
\end{tabular}
\end{table}

%\begin{figure}
%    \centering
%    \includegraphics[width=0.65\linewidth]{figures/cifar100c_f_clean_vs_corrupt.pdf}
%    \caption{Effect of gradient-norm scaling factor f on the clean accuracy vs. corruption accuracy trade-off on WideResNet/CIFAR-100.}
%    \label{fig:grad-norm-fac}
%\end{figure}

\subsubsection{Noise level}
\begin{figure*}[h]
    \centering
    \includegraphics[width=0.95\linewidth]{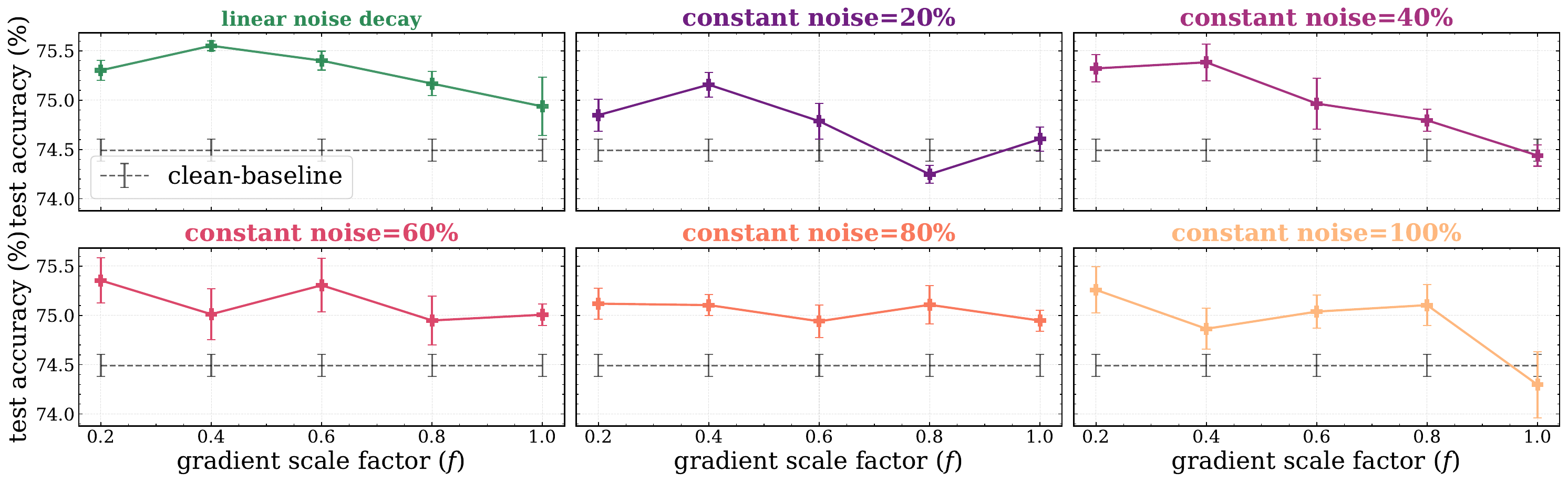}
    \caption{Ablation over gradient scaling factor f across noise levels on WideResNet/CIFAR-100, showing clean test accuracy as a function of the gradient-norm scaling factor $f$ for different noise levels $\sigma_t$. The top left panel shows the results of a linear decay of the noise level during an interleaved injection curriculum. The dashed line indicates the clean-only data baseline. }
    \label{fig:noise-ablation}
\end{figure*}

The amount of noise $\sigma_t$ added during each \textit{noisy} epoch is another hyperparameter. \autoref{fig:noise-ablation} shows an ablation over both the gradient-norm scaling factor and the level of noise injected during the noisy epochs. We also show a trial (in green) for linearly decaying the amount of noise injected. In general, we find a gradient-norm scaling factor of roughly $f\approx0.4$ performs best across all noise levels in terms of clean accuracy. We show a visual example of the noise levels when utilizing salt-and-pepper noise for CIFAR-100 samples in \autoref{fig:noise-grid}.

\subsubsection{Interleaved period}
\label{sec:ablation-p}

\begin{figure*}[h]
    \centering
    \includegraphics[width=0.95\linewidth]{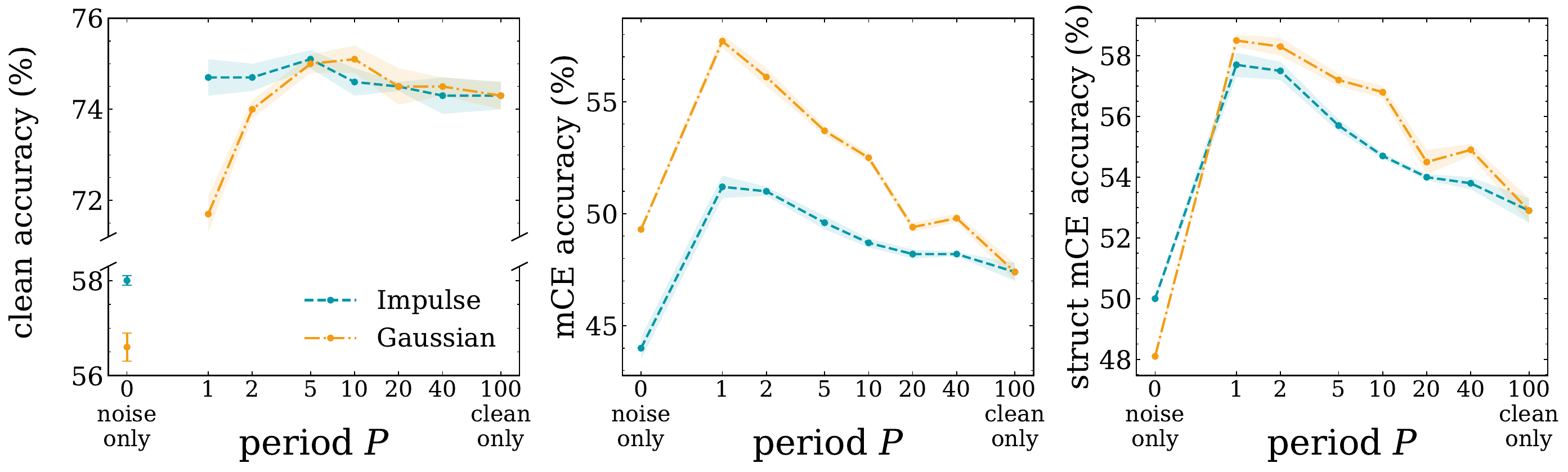}
    \caption{Ablation over the interleaved period P showing the same performance metrics with the same model/dataset as \autoref{tab:ablation_f}. Tabular results are provided in \autoref{tab:ablation_P}}
    \label{fig:period-ablation}
\end{figure*}

We also evaluate the sensitivity of the results to the interleaved period $P$, which determines the frequency of the noisy epochs (while leaving $L=1$ for simplicity). We perform trials again over the WideResNet on the CIFAR-100 dataset varying the period from 1 to 100 in \autoref{fig:period-ablation}. We find the best results occur for $P\in [5,10]$. Importantly, and as expected, lower values (more frequent noise) can harm clean accuracy but improve corruption performance, and conversely, larger values limit the regularization strength, thus reducing the corruption performance until noise injection has little to no effect on performance, regardless of noise distribution.

\section{Discussion}

\subsection{Interleaved curricula}
Our results demonstrate that noise injection, which has traditionally been invoked to improve optimization robustness, is more effective than previously known. It provides a tool to improve the performance under the optimization objective itself, but it has to be treated with nuance. In particular, the specific curriculum used to inject noise is as important as the noise distribution. Monotonic decay schedules lose their regularizing effect late in training, trapping the optimizer in local minima. Conversely, permanently applying heavy noise prevents the network from exploiting fine-grained features. We find the anti-curriculum loses the robustness benefit in the late training stages while standard curriculum fails to converge to accurate solutions on clean data (see \autoref{fig:curriculum}). Our proposed interleaved noise schedule resolves this conflict by mixing exploration during noisy epochs with exploitation during the clean epochs. Noise acts as a regularizer, forcing the optimizer out of sharp, overfitting minima, while the clean epochs allow the model to refine its decision boundaries without the interference from data distortion. Interleaving these periods prevents the model from catastrophically forgetting either of these regimes.

\subsection{Architecture-specific noise preferences}
\label{sec:architecture}
\begin{figure}[h]
    \centering
    \includegraphics[width=0.98\linewidth]{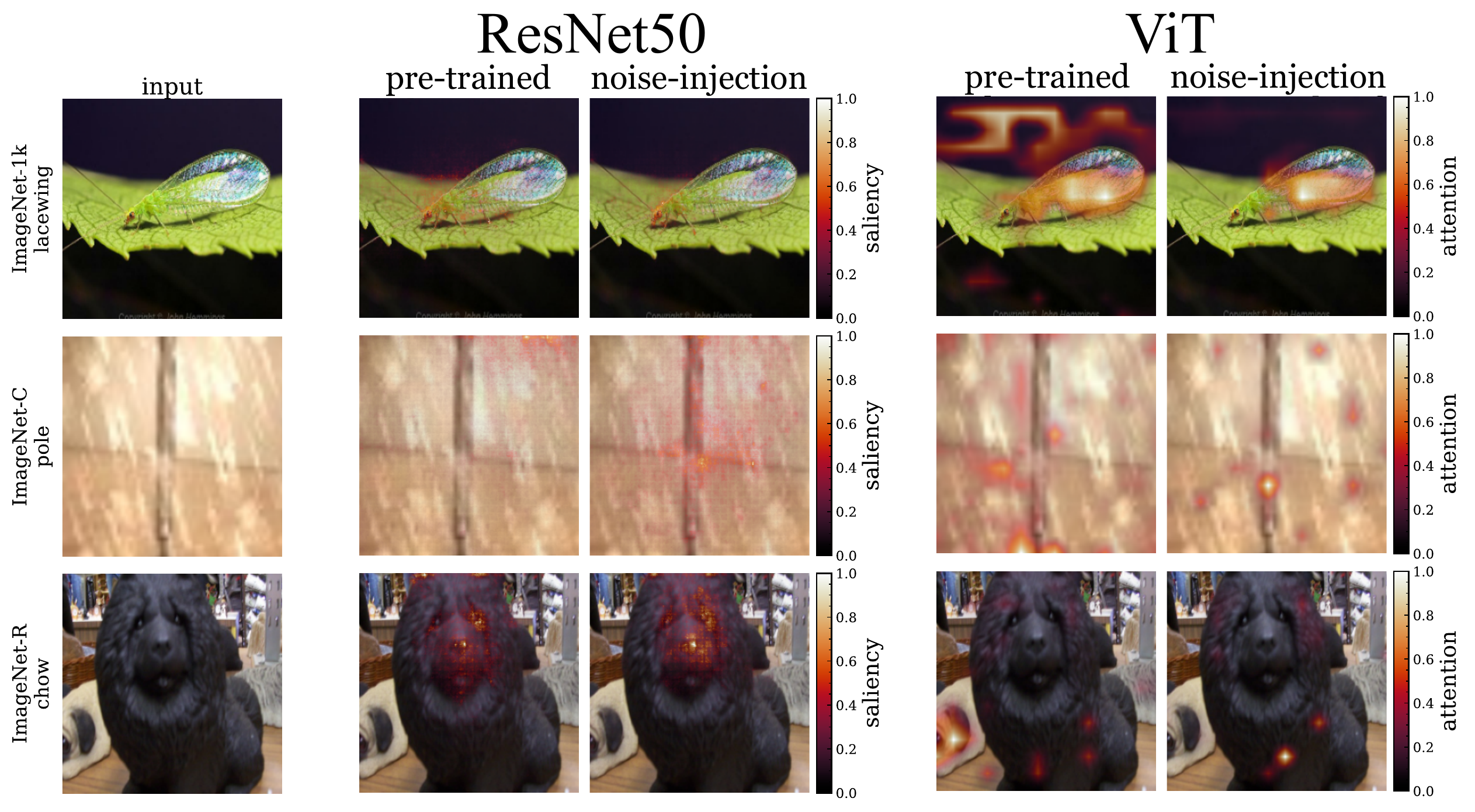}
    \caption{Visualization of the effect of interleaved noise injection on model focus for ResNet50 (left) and ViT (right). The rows show samples from ImageNet-1k (clean), ImageNet-C (corrupt) and ImageNet-R (OOD), with saliency and attention maps both before and after fine-tuning with noise injection. For the ResNet50, we calculate the gradient-based saliency maps; for the ViT, we show attention maps of the final layer (CLS) attention. After fine-tuning with interleaved noise, the ResNet50 saliency maps for the corrupt and OOD samples move away from high-frequency local features toward broader spatial regions, consistent with the Jacobian regularization effect derived in \autoref{eq:taylor}. The ViT attention maps show an opposite effect: Fine-tuning with interleaved Gaussian noise produces more uniform attention across patches and in particular, helps the model focus on the correct local features in all samples. }
    \label{fig:salient_attention}
\end{figure}

\autoref{tab:cifar_results} and \autoref{tab:imagenet_results} suggest that the choice of Gaussian vs impulse noise appears to be dependent on the underlying model architecture. The ResNet architectures work best with the impulse noise, while the vision transformer performs best the with Gaussian noise. 
We find the interaction between noise type and architecture intuitive in the context of the inductive biases of each architecture. The convolutional layers of the ResNet models create a strong locality bias. Impulse noise randomly corrupts individual pixel intensities, preventing the network from over-fitting to localized pixel-level features. Vision transformers, on the other hand, treat images as sequences of patches and thus lack this form of locality bias, instead learning primarily global features. Because Gaussian noise applies perturbations uniformly across all of these patches, it tends to smooth out low-contrast global features, which forces the attention mechanism to emphasize higher frequency, local features. Recent work by \citet{locality_vit} has demonstrated that explicitly enforcing locality structure in ViTs via architectural changes can significantly improve high frequency spatial representation, and we believe that applying interleaved Gaussian noise achieves precisely this training goal. 

\autoref{fig:salient_attention} shows a gradient saliency map for the ResNet50 and an attention map for the ViT, before and after fine-tuning with interleaved noise injection. %The top rows of this plot show the saliency/attention for a clean sample, with the bottom rows showing the same for a noisy sample, we plot the normalized difference between the pretrained and fine-tuned models in the rightmost column. 
We observe that the ResNet50 models initially concentrate heavily on a small number of important pixels, in clean and noisy samples. After the fine-tuning with our impulse noise curriculum, the focus is more diffuse and appears to capture a much larger area of the main feature, especially for the noisy sample (2nd row). Conversely, the pretrained ViT shows very little focus on the main feature, especially for the noisy sample, where the attention is drawn to weak, spurious features of the snow corruption. After the interleaved noise injection with Gaussian noise, the focus becomes significantly sharper with the majority of the attention being directed at the telephone pole. Interestingly, the same noise injection yields better generalization to OOD samples (3rd row) as well, as evidenced by a shifted focus to the dog in the middle (the actual chow-chow) instead of the other dog on the left edge of the image.

\subsection{Combination with alternative augmentations}
Despite being powerful on its own, another advantage of noise injection (interleaved or otherwise) lies in its ability to be combined, at virtually no extra cost, with existing training and augmentation pipelines. Our results demonstrate that the interleaved noise curriculum acts to further improve the performance of established robustness-enhancing training techniques, such as Cutout, AugMix, SAM, and VIPAug.

As SAM operates entirely in the weight space, it is not surprising that our method (which operates in data space) provides complementary benefits. But we have clearly seen that our method also improves the performance of spatial and frequency-space data augmentations like Cutout, AugMix, and VIPAug. These augmentations directly alter the structure of the input samples to encourage invariance under the respective equivalence transformations. So, why does a simple data-space noise injection provide additional benefits beyond these more complex data-space augmentations?

We believe the improvements from interleaved noise injection primarily come from altering the optimization dynamics through the regularizer. Specifically, from improving the ability to escape sub-optimal local minima of the loss surface. By injecting noise into the data, the gradient signal is randomized, causing a random walk with drift. The drift component stems from the original gradient, which is not fully destroyed by the noise. If the optimizer is stuck in a sharp minimum, the random component will allow it to escape, so that it ends in larger, flatter minima that are harder to leave through a random walk. Finding such flat minima has been found to lead to better overall solutions as well as increased generalization and robustness \citep{ly2025Nature}. The gradient-norm scaling is therefore a critical component of our method because it provides a balance between the random and the drift component.

\section{Conclusions}
In this study we introduce an interleaved noise injection scheme and demonstrate that periodically shifting between clean and noisy training epochs improves the test performance on clean data compared to the customary optimization that trains only with clean data. This benefit comes at essentially zero computational cost, does not require more training data or additional optimization epochs, and requires only minimal hyperparameter tuning. 

We provide a theoretical analysis to demonstrate that noise injection is equivalent to regularization. Specifically, we show that impulse noise introduces penalties that are proportional to the Jacobian and the Hessian of the loss with respect to the data; for Gaussian noise only a previously known curvature term remains. 

The benefits of these penalties can best be realized through an interleaved schedule compared to alternative curricula, which either transition only once from noisy to clean data or linearly change the noise level. We believe that repeatedly switching between noisy and clean data encourages exploration during the noisy epochs without entirely forgetting the features of the clean data.
This combination allows optimizers to escape sub-optimal local minima, reducing overfitting and improving performance on in-domain, out-of-domain, and corrupted samples.

We also find that the type of noise is best chosen to combat the failure modes of different model architectures. While interleaved noise injection generally improves training performance, convolutional architectures tend to benefit more from impulse noise, whereas attention-based models benefit more from Gaussian noise.

Lastly, we show that our method can also be combined with other common forms of corruption and augmentation to further improve their performance with minimal added computational overhead. 

Overall, interleaved noise injection provides a simple, yet effective tool to improve model performance and robustness against noise and real-world distribution shifts.

\section*{Software and Data}
We use \texttt{jax} \citep{jax2018github}, \texttt{pytorch}
\citep{paszke2019pytorch}, and \texttt{tensorflow} \citep{abadi2016tensorflow} for model architecture and data-loading. We use the publicly available CIFAR-100, CIFAR-100-C, ImageNet-1k, ImageNet-C, and ImageNet-R datasets \citep{hendrycks2019benchmarking,hendrycks2021many}. 

% Acknowledgements should only appear in the accepted version.
%\section*{Acknowledgements}

\section*{Impact Statement}
This paper presents work whose goal is to advance the field of Machine
Learning. There are many potential societal consequences of our work, none
of which we feel must be specifically highlighted here.

\bibliography{main}

@article{franzke_annealing,
  title = {Noise can speed Markov chain Monte Carlo estimation and quantum annealing},
  author = {Franzke, Brandon and Kosko, Bart},
  journal = {Phys. Rev. E},
  volume = {100},
  issue = {5},
  pages = {053309},
  numpages = {18},
  year = {2019},
  month = {Nov},
  publisher = {American Physical Society},
  doi = {10.1103/PhysRevE.100.053309},
  url = {https://link.aps.org/doi/10.1103/PhysRevE.100.053309}
}

@inproceedings{curriculum_bengio,
author = {Bengio, Yoshua and Louradour, J\'{e}r\^{o}me and Collobert, Ronan and Weston, Jason},
title = {Curriculum learning},
year = {2009},
isbn = {9781605585161},
publisher = {Association for Computing Machinery},
address = {New York, NY, USA},
url = {https://doi.org/10.1145/1553374.1553380},
doi = {10.1145/1553374.1553380},
booktitle = {Proceedings of the 26th Annual International Conference on Machine Learning},
pages = {41–48},
numpages = {8},
location = {Montreal, Quebec, Canada},
series = {ICML '09}
}

@article{hochreiter1994simplifying,
  title={Simplifying neural nets by discovering flat minima},
  author={Hochreiter, Sepp and Schmidhuber, J{\"u}rgen},
  journal={Advances in neural information processing systems},
  volume={7},
  year={1994}
}

@article{hochreiter1997flat,
  title={Flat minima},
  author={Hochreiter, Sepp and Schmidhuber, J{\"u}rgen},
  journal={Neural computation},
  volume={9},
  number={1},
  pages={1--42},
  year={1997},
  publisher={MIT Press One Rogers Street, Cambridge, MA 02142-1209, USA journals-info~…}
}

@inproceedings{hardt2016train,
  title={Train faster, generalize better: Stability of stochastic gradient descent},
  author={Hardt, Moritz and Recht, Ben and Singer, Yoram},
  booktitle={International conference on machine learning},
  pages={1225--1234},
  year={2016},
  organization={PMLR}
}

@inproceedings{xie2021positive,
  title={Positive-negative momentum: Manipulating stochastic gradient noise to improve generalization},
  author={Xie, Zeke and Yuan, Li and Zhu, Zhanxing and Sugiyama, Masashi},
  booktitle={International Conference on Machine Learning},
  pages={11448--11458},
  year={2021},
  organization={PMLR}
}

@article{wu2020direction,
  title={Direction matters: On the implicit bias of stochastic gradient descent with moderate learning rate},
  author={Wu, Jingfeng and Zou, Difan and Braverman, Vladimir and Gu, Quanquan},
  journal={arXiv preprint arXiv:2011.02538},
  year={2020}
}

@inproceedings{battash2024revisiting,
  title={Revisiting the noise model of stochastic gradient descent},
  author={Battash, Barak and Wolf, Lior and Lindenbaum, Ofir},
  booktitle={International Conference on Artificial Intelligence and Statistics},
  pages={4780--4788},
  year={2024},
  organization={PMLR}
}

@article{marinari1992simulated,
  title={Simulated tempering: a new Monte Carlo scheme},
  author={Marinari, Enzo and Parisi, Giorgio},
  journal={Europhysics letters},
  volume={19},
  number={6},
  pages={451},
  year={1992},
  publisher={IOP Publishing}
}

@article{earl2005parallel,
  title={Parallel tempering: Theory, applications, and new perspectives},
  author={Earl, David J and Deem, Michael W},
  journal={Physical Chemistry Chemical Physics},
  volume={7},
  number={23},
  pages={3910--3916},
  year={2005},
  publisher={Royal Society of Chemistry}
}

@article{hoffman2019robust,
  title={Robust learning with jacobian regularization},
  author={Hoffman, Judy and Roberts, Daniel A and Yaida, Sho},
  journal={arXiv preprint arXiv:1908.02729},
  year={2019}
}

@article{chan2019jacobian,
  title={Jacobian adversarially regularized networks for robustness},
  author={Chan, Alvin and Tay, Yi and Ong, Yew Soon and Fu, Jie},
  journal={arXiv preprint arXiv:1912.10185},
  year={2019}
}

@inproceedings{jakubovitz2018improving,
  title={Improving dnn robustness to adversarial attacks using jacobian regularization},
  author={Jakubovitz, Daniel and Giryes, Raja},
  booktitle={Proceedings of the European conference on computer vision (ECCV)},
  pages={514--529},
  year={2018}
}

@article{hendrycks2019benchmarking,
  title={Benchmarking neural network robustness to common corruptions and perturbations},
  author={Hendrycks, Dan and Dietterich, Thomas},
  journal={arXiv preprint arXiv:1903.12261},
  year={2019}
}

@inproceedings{hendrycks2021many,
  title={The many faces of robustness: A critical analysis of out-of-distribution generalization},
  author={Hendrycks, Dan and Basart, Steven and Mu, Norman and Kadavath, Saurav and Wang, Frank and Dorundo, Evan and Desai, Rahul and Zhu, Tyler and Parajuli, Samyak and Guo, Mike and others},
  booktitle={Proceedings of the IEEE/CVF international conference on computer vision},
  pages={8340--8349},
  year={2021}
}

@article{foret2020sharpness,
  title={Sharpness-aware minimization for efficiently improving generalization},
  author={Foret, Pierre and Kleiner, Ariel and Mobahi, Hossein and Neyshabur, Behnam},
  journal={arXiv preprint arXiv:2010.01412},
  year={2020}
}

@inproceedings{cohen2019certified,
  title={Certified adversarial robustness via randomized smoothing},
  author={Cohen, Jeremy and Rosenfeld, Elan and Kolter, Zico},
  booktitle={international conference on machine learning},
  pages={1310--1320},
  year={2019},
  organization={PMLR}
}

@article{hendrycks2019augmix,
  title={Augmix: A simple data processing method to improve robustness and uncertainty},
  author={Hendrycks, Dan and Mu, Norman and Cubuk, Ekin D and Zoph, Barret and Gilmer, Justin and Lakshminarayanan, Balaji},
  journal={arXiv preprint arXiv:1912.02781},
  year={2019}
}

@inproceedings{he2016deep,
  title={Deep residual learning for image recognition},
  author={He, Kaiming and Zhang, Xiangyu and Ren, Shaoqing and Sun, Jian},
  booktitle={Proceedings of the IEEE conference on computer vision and pattern recognition},
  pages={770--778},
  year={2016}
}

@article{dosovitskiy2020image,
  title={An image is worth 16x16 words: Transformers for image recognition at scale},
  author={Dosovitskiy, Alexey and Beyer, Lucas and Kolesnikov, Alexander and Weissenborn, Dirk and Zhai, Xiaohua and Unterthiner, Thomas and Dehghani, Mostafa and Minderer, Matthias and Heigold, Georg and Gelly, Sylvain and others},
  journal={arXiv preprint arXiv:2010.11929},
  year={2020}
}

@article{zagoruyko2016wide,
  title={Wide residual networks},
  author={Zagoruyko, Sergey and Komodakis, Nikos},
  journal={arXiv preprint arXiv:1605.07146},
  year={2016}
}

@article{devries2017improved,
  title={Improved regularization of convolutional neural networks with cutout},
  author={DeVries, Terrance and Taylor, Graham W},
  journal={arXiv preprint arXiv:1708.04552},
  year={2017}
}

@inproceedings{lee2024domain,
  title={Domain generalization with vital phase augmentation},
  author={Lee, Ingyun and Lee, Wooju and Myung, Hyun},
  booktitle={Proceedings of the AAAI conference on artificial intelligence},
  volume={38},
  number={4},
  pages={2892--2900},
  year={2024}
}

@article{rohrer2015interleaved,
  title={Interleaved practice improves mathematics learning.},
  author={Rohrer, Doug and Dedrick, Robert F and Stershic, Sandra},
  journal={Journal of Educational Psychology},
  volume={107},
  number={3},
  pages={900},
  year={2015},
  publisher={American Psychological Association}
}

@article{taylor2010effects,
  title={The effects of interleaved practice},
  author={Taylor, Kelli and Rohrer, Doug},
  journal={Applied cognitive psychology},
  volume={24},
  number={6},
  pages={837--848},
  year={2010},
  publisher={Wiley Online Library}
}

@article{firth2021systematic,
  title={A systematic review of interleaving as a concept learning strategy},
  author={Firth, Jonathan and Rivers, Ian and Boyle, James},
  journal={Review of Education},
  volume={9},
  number={2},
  pages={642--684},
  year={2021},
  publisher={Wiley Online Library}
}

@software{jax2018github,
  author = {James Bradbury and Roy Frostig and Peter Hawkins and Matthew James Johnson and Chris Leary and Dougal Maclaurin and George Necula and Adam Paszke and Jake Vander{P}las and Skye Wanderman-{M}ilne and Qiao Zhang},
  title = {{JAX}: composable transformations of {P}ython+{N}um{P}y programs},
  url = {http://github.com/jax-ml/jax},
  version = {0.3.13},
  year = {2018},
}

@article{paszke2019pytorch,
  title={Pytorch: An imperative style, high-performance deep learning library},
  author={Paszke, Adam and Gross, Sam and Massa, Francisco and Lerer, Adam and Bradbury, James and Chanan, Gregory and Killeen, Trevor and Lin, Zeming and Gimelshein, Natalia and Antiga, Luca and others},
  journal={Advances in neural information processing systems},
  volume={32},
  year={2019}
}

@inproceedings{abadi2016tensorflow,
  title={$\{$TensorFlow$\}$: a system for $\{$Large-Scale$\}$ machine learning},
  author={Abadi, Mart{\'\i}n and Barham, Paul and Chen, Jianmin and Chen, Zhifeng and Davis, Andy and Dean, Jeffrey and Devin, Matthieu and Ghemawat, Sanjay and Irving, Geoffrey and Isard, Michael and others},
  booktitle={12th USENIX symposium on operating systems design and implementation (OSDI 16)},
  pages={265--283},
  year={2016}
}

@inproceedings{he2019parametric,
  title={Parametric noise injection: Trainable randomness to improve deep neural network robustness against adversarial attack},
  author={He, Zhezhi and Rakin, Adnan Siraj and Fan, Deliang},
  booktitle={Proceedings of the IEEE/CVF conference on computer vision and pattern recognition},
  pages={588--597},
  year={2019}
}

@article{xiao2024noise,
  title={Noise optimization in artificial neural networks},
  author={Xiao, Li and Zhang, Zeliang and Huang, Kuihua and Jiang, Jinyang and Peng, Yijie},
  journal={IEEE Transactions on Automation Science and Engineering},
  volume={22},
  pages={2780--2793},
  year={2024},
  publisher={IEEE}
}

@article{grandvalet1997noise,
  title={Noise injection: Theoretical prospects},
  author={Grandvalet, Yves and Canu, St{\'e}phane and Boucheron, St{\'e}phane},
  journal={Neural Computation},
  volume={9},
  number={5},
  pages={1093--1108},
  year={1997},
  publisher={MIT Press}
}

@article{zur2009noise,
  title={Noise injection for training artificial neural networks: A comparison with weight decay and early stopping},
  author={Zur, Richard M and Jiang, Yulei and Pesce, Lorenzo L and Drukker, Karen},
  journal={Medical physics},
  volume={36},
  number={10},
  pages={4810--4818},
  year={2009},
  publisher={Wiley Online Library}
}

@article{matsuoka1992noise,
  title={Noise injection into inputs in back-propagation learning},
  author={Matsuoka, Kiyotoshi},
  journal={IEEE Transactions on Systems, Man, and Cybernetics},
  volume={22},
  number={3},
  pages={436--440},
  year={1992},
  publisher={IEEE}
}

@ARTICLE{locality_vit,
       author = {{Hajimiri}, Sina and {Beizaee}, Farzad and {Shakeri}, Fereshteh and {Desrosiers}, Christian and {Ben Ayed}, Ismail and {Dolz}, Jose},
        title = "{Locality-Attending Vision Transformer}",
      journal = {arXiv e-prints},
         year = 2026,
        month = mar,
          eid = {arXiv:2603.04892},
        pages = {arXiv:2603.04892},
archivePrefix = {arXiv},
       eprint = {2603.04892},
 primaryClass = {cs.CV},
       adsurl = {https://ui.adsabs.harvard.edu/abs/2026arXiv260304892H}
}

@article{levi2022noise,
  title={Noise injection node regularization for robust learning},
  author={Levi, Noam and Bloch, Itay M and Freytsis, Marat and Volansky, Tomer},
  journal={arXiv preprint arXiv:2210.15764},
  year={2022}
}

@article{gan2026neural,
  title={Neural Thickets: Diverse Task Experts Are Dense Around Pretrained Weights},
  author={Gan, Yulu and Isola, Phillip},
  journal={arXiv preprint arXiv:2603.12228},
  year={2026}
}

@ARTICLE{ly2025Nature,
       author = {{Ly}, Andrew and {Gong}, Pulin},
        title = "{Optimization on multifractal loss landscapes explains a diverse range of geometrical and dynamical properties of deep learning}",
      journal = {Nature Communications},
         year = 2025,
        month = apr,
       volume = {16},
       number = {1},
          eid = {3252},
        pages = {3252},
          doi = {10.1038/s41467-025-58532-9},
       adsurl = {https://ui.adsabs.harvard.edu/abs/2025NatCo..16.3252L}
}

@ARTICLE{spiking_noise,
       author = {{Zhang}, Zeliang and {Jiang}, Jinyang and {Chen}, Minjie and {Wang}, Zhiyuan and {Peng}, Yijie and {Yu}, Zhaofei},
        title = "{A Novel Noise Injection-based Training Scheme for Better Model Robustness}",
      journal = {arXiv e-prints},
         year = 2023,
        month = feb,
          eid = {arXiv:2302.10802},
        pages = {arXiv:2302.10802},
          doi = {10.48550/arXiv.2302.10802},
archivePrefix = {arXiv},
       eprint = {2302.10802},
 primaryClass = {cs.LG},
       adsurl = {https://ui.adsabs.harvard.edu/abs/2023arXiv230210802Z}
}
\bibliographystyle{tmlr}

%%%%%%%%%%%%%%%%%%%%%%%%%%%%%%%%%%%%%%%%%%%%%%%%%%%%%%%%%%%%%%%%%%%%%%%%%%%%%%%
%%%%%%%%%%%%%%%%%%%%%%%%%%%%%%%%%%%%%%%%%%%%%%%%%%%%%%%%%%%%%%%%%%%%%%%%%%%%%%%
% APPENDIX
%%%%%%%%%%%%%%%%%%%%%%%%%%%%%%%%%%%%%%%%%%%%%%%%%%%%%%%%%%%%%%%%%%%%%%%%%%%%%%%
%%%%%%%%%%%%%%%%%%%%%%%%%%%%%%%%%%%%%%%%%%%%%%%%%%%%%%%%%%%%%%%%%%%%%%%%%%%%%%%
\newpage
\appendix
\onecolumn

\section{Effect of gradient-norm stabilization}
We find $f= 0.4$ to provide the best results in our tests, i.e., we take steps a bit less than half the magnitude taken during the clean epochs, but we find any value between 0.2 and 0.8 to provide similar results. We present an ablation over the scale-factor $f$ in \autoref{sec:ablations}. Setting $f > 1$ means that during a noisy epoch, the optimizer will potentially move far away from desirable regions of the loss surface. Setting $f$ too small simply means that very small updates are performed during the noisy epochs, hence the benefits of our method vanish.

\begin{figure}[h]
    \centering
    \includegraphics[width=0.96\linewidth]{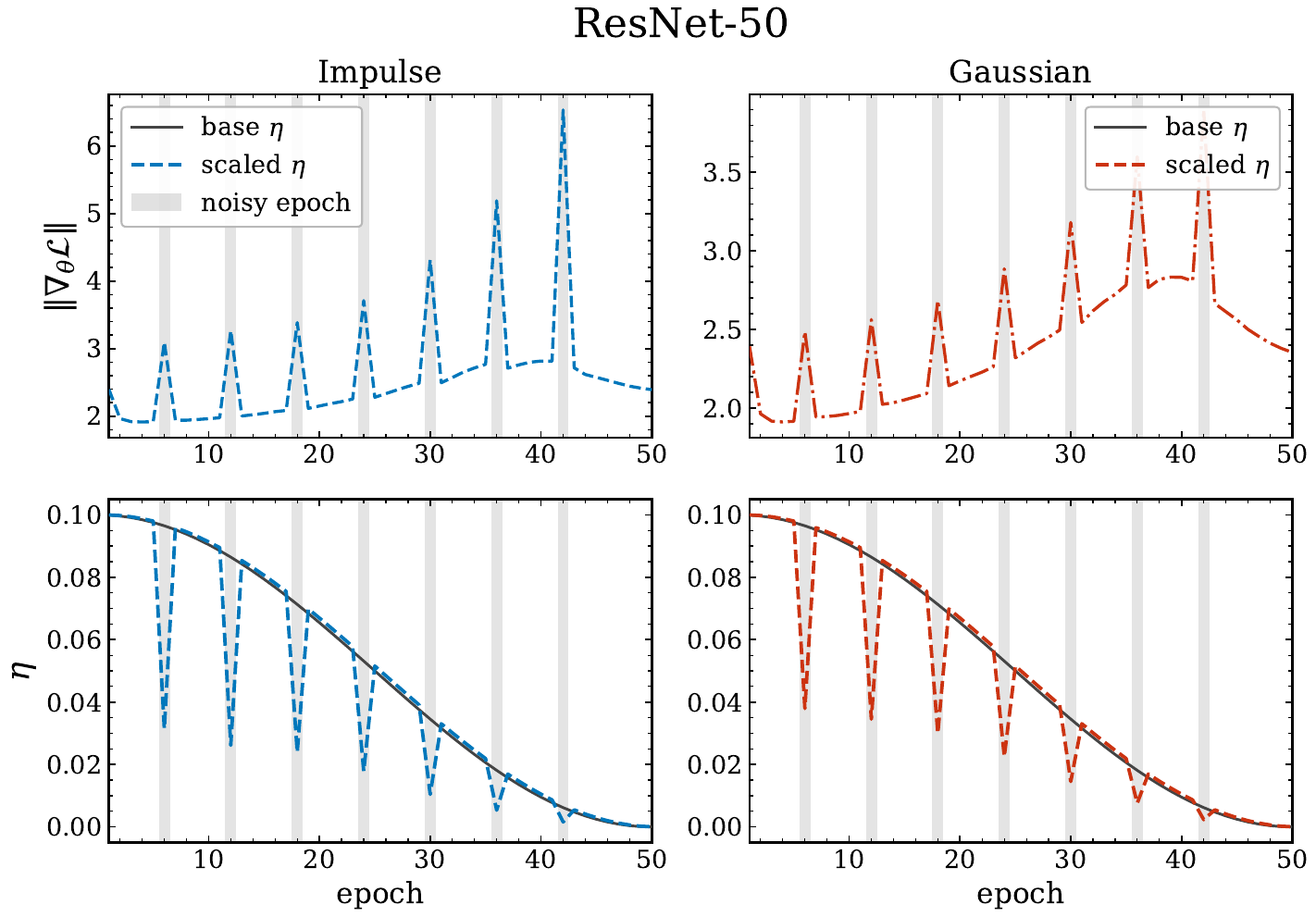}
    \caption{In the top row we show the magnitude of the gradient-norm before stabilization for the ResNet50 with interleaved noise using impulse and Gaussian noise from left to right respectively. The gray shading indicates noise has been injected for that epoch. In the bottom row we show the learning rate, and effective learning rate (after gradient-norm stabilization from \autoref{eq:stabalisation} with $f=0.4$).  }
    \label{fig:gradnorm-resnet}
\end{figure}

In \autoref{fig:gradnorm-resnet} we show the effect on the magnitude of the gradient norm, and hence the change to the effective learning rate via \autoref{eq:stabalisation}. We show this for a ResNet50 fine-tuned with interleaved noise injection on ImageNet-1k with impulse noise in the left column and Gaussian noise in the right. Note that further details on all trials and architecture setups are in the following section. We color the noisy epochs in gray shading, and see that this corresponds to strong spikes in $\| \nabla_{\theta} \mathcal{L}\|$ which can have strong destabilizing effects if not scaled down particularly in late training. We apply \autoref{eq:stabalisation} to the base $\eta$ to control the effective total update magnitude via using the new scaling $\eta$ as shown in the colored dashed lines in the top rows of \autoref{fig:gradnorm-resnet}.  We observe similar spiking when using a ViT for ImageNet-1k, and a WideResNet on CIFAR-100 which we show in \autoref{fig:gradnorm-vit} and \autoref{fig:gradnorm-wide} in the appendix.

\begin{figure}[h]
    \centering
    \includegraphics[width=0.92\linewidth]{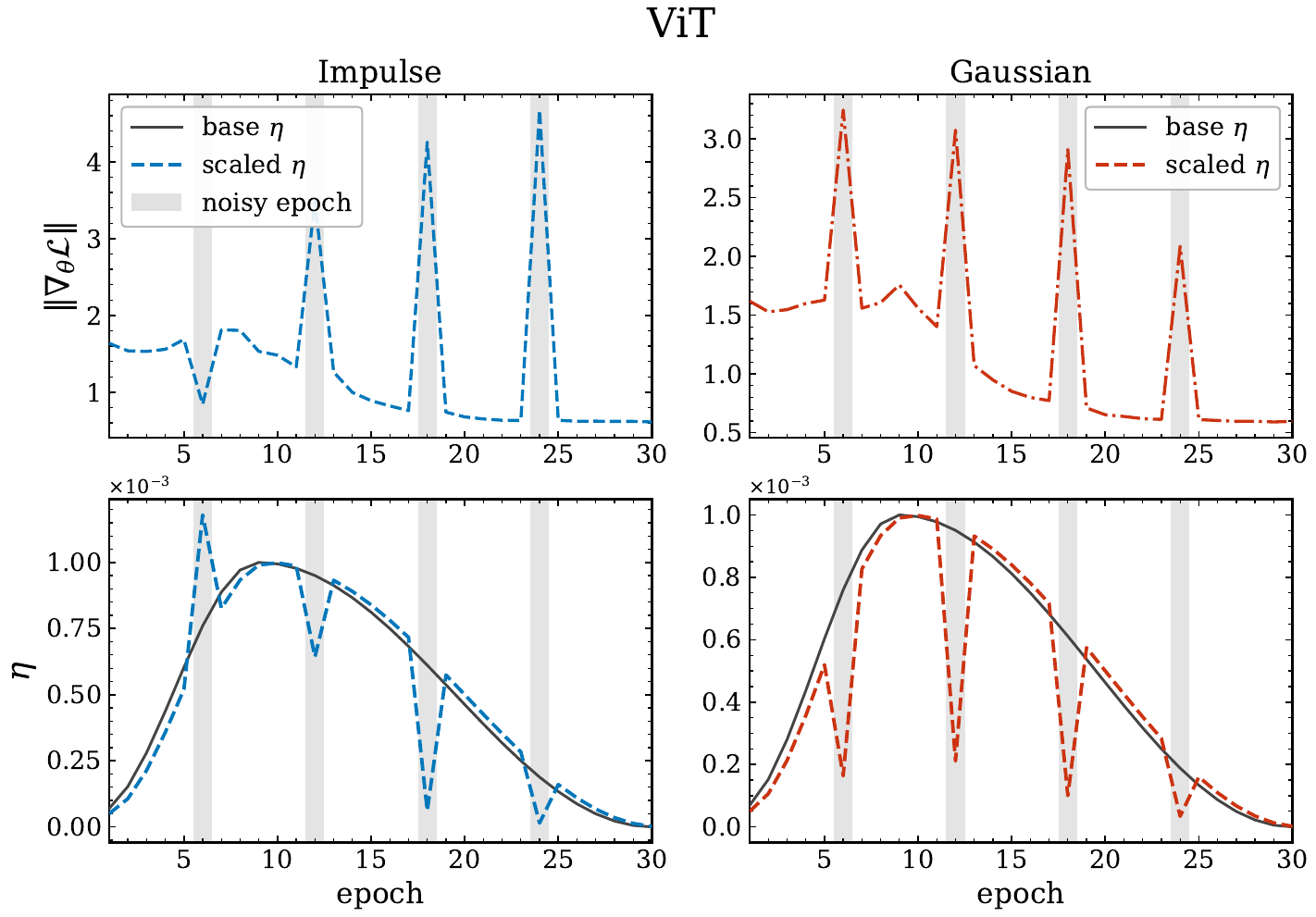}
    \caption{Same as \autoref{fig:gradnorm-resnet} but for the ViT architecture.}
    \label{fig:gradnorm-vit}
\end{figure}
\clearpage

\begin{figure}[h]
    \centering
    \includegraphics[width=0.92\linewidth]{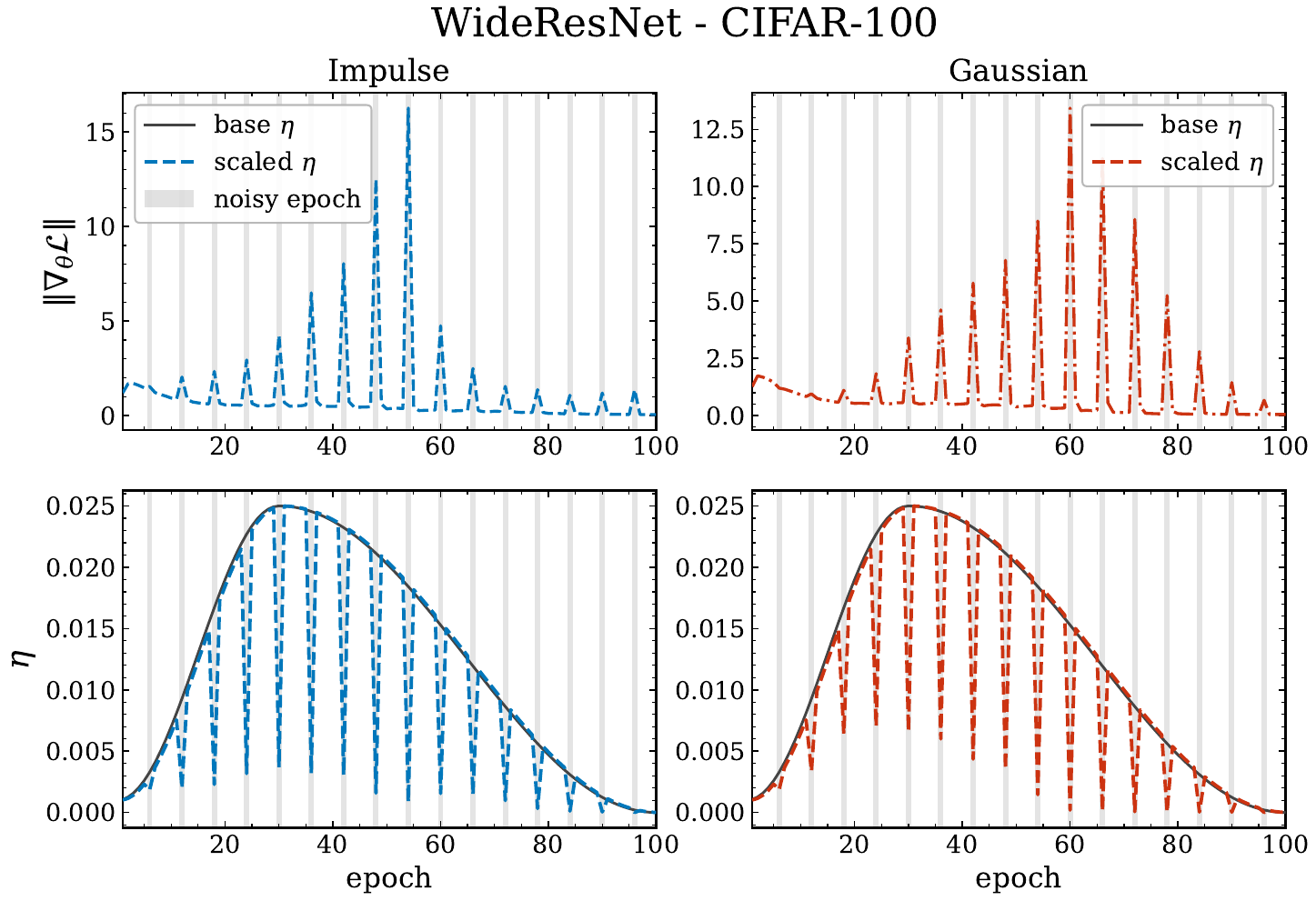}
    \caption{Same as \autoref{fig:gradnorm-resnet} but for the WideResNet of CIFAR-100.}
    \label{fig:gradnorm-wide}
\end{figure}

\section{Performance with corruption severity}
\label{sec:performance-severity}
%---------------------------
% by severity imagenet table
%\input{imagenet_severity_table}
% ------------------------------ TABLE 4 ------------------------------
\begin{sidewaystable}[p]
\caption{Comparison of baseline regularizers and interleaved noise augmentations
across corruption severity levels. Clean Error and Mean Corruption Error (mCE)
are reported alongside error rates for severities 1 through 5. All metrics are
error percentages ($\downarrow$), with bootstrapped standard errors.
Improvements over the respective baseline are shown in (green), regressions in (red).}
\label{tab:table4}
\begin{center}
\scriptsize
\setlength{\tabcolsep}{3pt}
\renewcommand{\arraystretch}{1.25} 
\begin{tabular}{ll lllllll}
\toprule
Architecture & Method & Clean Err ($\downarrow$) & mCE ($\downarrow$) & Sev 1 ($\downarrow$) & Sev 2 ($\downarrow$) & Sev 3 ($\downarrow$) & Sev 4 ($\downarrow$) & Sev 5 ($\downarrow$)\\
\midrule
\multirow{15}{*}{ResNet50}
& SGD               & $25.8\pm.1$ & $89.1\pm.2$ & $82.6\pm.4$ & $86.1\pm.3$ & $89.4\pm.3$ & $92.6\pm.2$ & $94.7\pm.1$ \\
& SGD + Gaus.       & $25.5\pm.2$ \improv{-0.3} & $88.2\pm.2$ \improv{-0.9} & $81.4\pm.1$ \improv{-1.2} & $85.1\pm.2$ \improv{-1.0} & $88.5\pm.2$ \improv{-0.9} & $91.9\pm.2$ \improv{-0.7} & $94.2\pm.2$ \improv{-0.5} \\
& SGD + Impulse     & $25.1\pm.2$ \improv{-0.7} & $87.3\pm.1$ \improv{-1.8} & $80.2\pm.4$ \improv{-2.4} & $84.0\pm.2$ \improv{-2.1} & $87.6\pm.1$ \improv{-1.8} & $91.1\pm.1$ \improv{-1.5} & $93.5\pm.1$ \improv{-1.2} \\
\cmidrule{2-9}
& Cutout            & $24.4\pm.1$ & $88.2\pm.3$ & $81.2\pm.6$ & $85.0\pm.5$ & $88.5\pm.5$ & $91.9\pm.4$ & $94.3\pm.3$ \\
& Cutout + Gaus.    & $24.5\pm.1$ \worse{+0.1}  & $88.0\pm.2$ \improv{-0.2} & $80.9\pm.3$ \improv{-0.3} & $84.8\pm.4$ \improv{-0.2} & $88.4\pm.4$ \improv{-0.1} & $91.9\pm.3$ \improv{-0.0} & $94.2\pm.2$ \improv{-0.1} \\
& Cutout + Impulse  & $23.7\pm.1$ \improv{-0.7} & $86.5\pm.3$ \improv{-1.7} & $78.6\pm.5$ \improv{-2.6} & $82.6\pm.6$ \improv{-2.4} & $86.5\pm.5$ \improv{-2.0} & $90.4\pm.5$ \improv{-1.5} & $93.1\pm.4$ \improv{-1.2} \\
\cmidrule{2-9}
& SAM               & $24.9\pm.2$ & $88.3\pm.2$ & $81.6\pm.5$ & $85.2\pm.5$ & $88.6\pm.3$ & $91.9\pm.2$ & $94.2\pm.2$ \\
& SAM + Gaus.       & $24.7\pm.1$ \improv{-0.2} & $85.7\pm.1$ \improv{-2.6} & $80.2\pm.2$ \improv{-1.4} & $84.1\pm.2$ \improv{-1.1} & $87.7\pm.1$ \improv{-0.9} & $91.4\pm.2$ \improv{-0.5} & $93.8\pm.2$ \improv{-0.4} \\
& SAM + Impulse     & $24.8\pm.1$ \improv{-0.1} & $84.8\pm.1$ \improv{-3.5} & $78.1\pm.3$ \improv{-3.5} & $82.1\pm.2$ \improv{-3.1} & $86.1\pm.2$ \improv{-2.5} & $89.9\pm.2$ \improv{-2.0} & $92.7\pm.1$ \improv{-1.5} \\
\cmidrule{2-9}
& VIPAug            & $28.7\pm.1$ & $89.7\pm.2$ & $83.4\pm.6$ & $87.0\pm.5$ & $90.1\pm.4$ & $93.0\pm.2$ & $95.0\pm.1$ \\
& VIPAug + Gaus.    & $28.6\pm.3$ \improv{-0.1} & $89.6\pm.3$ \improv{-0.1} & $83.0\pm.6$ \improv{-0.4} & $86.7\pm.6$ \improv{-0.3} & $90.1\pm.5$ \improv{-0.0} & $93.1\pm.4$ \worse{+0.1}  & $95.0\pm.3$ \improv{-0.0} \\
& VIPAug + Impulse  & $28.1\pm.3$ \improv{-0.6} & $88.6\pm.2$ \improv{-1.1} & $80.9\pm.3$ \improv{-2.5} & $84.7\pm.4$ \improv{-2.3} & $88.3\pm.2$ \improv{-1.8} & $91.6\pm.2$ \improv{-1.4} & $93.9\pm.1$ \improv{-1.1} \\
\cmidrule{2-9}
& AugMix            & $23.8\pm.2$ & $83.2\pm.2$ & $75.1\pm.3$ & $79.0\pm.3$ & $83.2\pm.3$ & $87.7\pm.3$ & $91.0\pm.2$ \\
& AugMix + Gaus.    & $23.7\pm.2$ \improv{-0.1} & $82.3\pm.1$ \improv{-0.9} & $74.1\pm.1$ \improv{-1.0} & $78.0\pm.1$ \improv{-1.0} & $82.2\pm.2$ \improv{-1.0} & $87.1\pm.2$ \improv{-0.6} & $90.5\pm.2$ \improv{-0.5} \\
& AugMix + Impulse  & $23.3\pm.2$ \improv{-0.5} & $82.0\pm.1$ \improv{-1.2} & $73.8\pm.1$ \improv{-1.3} & $77.7\pm.1$ \improv{-1.3} & $81.9\pm.1$ \improv{-1.3} & $86.6\pm.1$ \improv{-1.1} & $90.1\pm.1$ \improv{-0.9} \\
\midrule \midrule
\multirow{15}{*}{ViT}
& AdamW             & $20.1\pm.3$ & $69.5\pm.4$ & $57.1\pm.7$ & $62.4\pm.7$ & $69.2\pm.7$ & $76.5\pm.6$ & $82.1\pm.6$ \\
& AdamW + Gaus.     & $19.5\pm.3$ \improv{-0.6} & $66.1\pm.1$ \improv{-3.4} & $53.7\pm.2$ \improv{-3.4} & $58.8\pm.1$ \improv{-3.6} & $65.5\pm.1$ \improv{-3.7} & $73.2\pm.2$ \improv{-3.3} & $79.6\pm.2$ \improv{-2.5} \\
& AdamW + Impulse   & $20.0\pm.3$ \improv{-0.1} & $67.7\pm.3$ \improv{-1.8} & $56.1\pm.9$ \improv{-1.0} & $60.7\pm.6$ \improv{-1.7} & $67.1\pm.6$ \improv{-2.1} & $74.3\pm.4$ \improv{-2.2} & $80.0\pm.4$ \improv{-2.1} \\
\cmidrule{2-9}
& Cutout            & $19.9\pm.1$ & $69.4\pm.4$ & $57.0\pm.3$ & $62.3\pm.4$ & $69.2\pm.4$ & $76.5\pm.4$ & $82.2\pm.4$ \\
& Cutout + Gaus.    & $19.2\pm.1$ \improv{-0.7} & $66.3\pm.2$ \improv{-3.1} & $54.1\pm.5$ \improv{-2.9} & $59.0\pm.4$ \improv{-3.3} & $65.6\pm.6$ \improv{-3.6} & $73.2\pm.5$ \improv{-3.3} & $79.6\pm.5$ \improv{-2.6} \\
& Cutout + Impulse  & $20.3\pm.1$ \worse{+0.4}  & $68.8\pm.4$ \improv{-0.6} & $56.9\pm.5$ \improv{-0.1} & $61.8\pm.7$ \improv{-0.5} & $68.5\pm.7$ \improv{-0.7} & $75.6\pm.7$ \improv{-0.9} & $81.2\pm.8$ \improv{-1.0} \\
\cmidrule{2-9}
& SAM               & $17.5\pm.1$ & $63.9\pm.2$ & $50.6\pm.2$ & $55.9\pm.2$ & $63.1\pm.1$ & $71.6\pm.1$ & $78.5\pm.1$ \\
& SAM + Gaus.       & $17.4\pm.1$ \improv{-0.1} & $63.1\pm.1$ \improv{-0.8} & $50.1\pm.2$ \improv{-0.5} & $55.3\pm.3$ \improv{-0.6} & $62.2\pm.2$ \improv{-0.9} & $70.5\pm.1$ \improv{-1.1} & $77.4\pm.1$ \improv{-1.1} \\
& SAM + Impulse     & $17.7\pm.1$ \worse{+0.2}  & $63.6\pm.1$ \improv{-0.3} & $51.1\pm.4$ \worse{+0.5}  & $55.9\pm.4$ \improv{-0.0} & $62.6\pm.4$ \improv{-0.5} & $70.5\pm.5$ \improv{-1.1} & $77.2\pm.5$ \improv{-1.3} \\
\cmidrule{2-9}
& VIPAug            & $19.8\pm.2$ & $69.8\pm.6$ & $56.8\pm.6$ & $62.2\pm.3$ & $69.2\pm.3$ & $76.5\pm.2$ & $81.9\pm.2$ \\
& VIPAug + Gaus.    & $18.9\pm.1$ \improv{-0.9} & $66.2\pm.2$ \improv{-3.6} & $53.8\pm.2$ \improv{-3.0} & $58.9\pm.2$ \improv{-3.3} & $65.5\pm.3$ \improv{-3.7} & $73.3\pm.3$ \improv{-3.2} & $79.6\pm.3$ \improv{-2.3} \\
& VIPAug + Impulse  & $20.0\pm.1$ \worse{+0.2}  & $68.7\pm.5$ \improv{-1.1} & $55.8\pm.9$ \improv{-1.0} & $60.6\pm.9$ \improv{-1.6} & $67.1\pm.9$ \improv{-2.1} & $74.5\pm.8$ \improv{-2.0} & $80.2\pm.7$ \improv{-1.7} \\
\cmidrule{2-9}
& AugMix            & $17.8\pm.1$ & $62.4\pm.1$ & $49.9\pm.2$ & $54.5\pm.1$ & $61.2\pm.2$ & $69.5\pm.1$ & $76.8\pm.2$ \\
& AugMix + Gaus.    & $17.5\pm.1$ \improv{-0.3} & $60.8\pm.1$ \improv{-1.6} & $49.1\pm.3$ \improv{-0.8} & $53.2\pm.2$ \improv{-1.3} & $59.4\pm.3$ \improv{-1.8} & $67.4\pm.4$ \improv{-2.1} & $74.7\pm.4$ \improv{-2.1} \\
& AugMix + Impulse  & $17.5\pm.1$ \improv{-0.3} & $61.6\pm.1$ \improv{-0.8} & $49.4\pm.1$ \improv{-0.5} & $53.8\pm.1$ \improv{-0.7} & $60.3\pm.1$ \improv{-0.9} & $68.5\pm.1$ \improv{-1.0} & $75.8\pm.1$ \improv{-1.0} \\
\bottomrule
\end{tabular}
\end{center}
\end{sidewaystable}

In \autoref{tab:table4} we show the performance of the ResNet50 and ViT models with the suite of robustness regularizers with the range of 5 corruption severities. We see that, as expected, the error rate increases with severity.

In \autoref{tab:imagenet_corruptions} we show the same results but by corruption type which is explained in detail in \citet{hendrycks2019augmix}.

%----------------------------
% by corruption type accuracy
\begin{sidewaystable}
\centering
\caption{Per-type corruption error rate (\%) on ImageNet-C averaged over severity levels for ResNet50 and ViT architectures.}
\label{tab:imagenet_corruptions}
\resizebox{\textwidth}{!}{
\begin{tabular}{l ccc cccc cccc cccc}
\toprule
& \multicolumn{3}{c}{\textbf{Noise}} & \multicolumn{4}{c}{\textbf{Blur}} & \multicolumn{4}{c}{\textbf{Weather}} & \multicolumn{4}{c}{\textbf{Digital}} \\
\cmidrule(lr){2-4} \cmidrule(lr){5-8} \cmidrule(lr){9-12} \cmidrule(lr){13-16}
\textbf{Method} & Gauss. & Shot & Imp. & Defoc. & Glass & Motion & Zoom & Snow & Frost & Fog & Bright. & Contr. & Elast. & Pixel. & JPEG \\
\midrule
\multicolumn{16}{c}{\textbf{ResNet50}} \\
\midrule
SGD & $93.5 \pm.3$ & $91.9 \pm.4$ & $93.5 \pm.4$ & $89.7 \pm.1$ & $95.6 \pm.2$ & $88.2 \pm.3$ & $88.3 \pm.2$ & $87.8 \pm.1$ & $86.4 \pm.3$ & $87.3 \pm.2$ & $79.1 \pm.2$ & $95.1 \pm.1$ & $86.3 \pm.3$ & $86.9 \pm.3$ & $86.5 \pm.4$ \\
SGD + Gaussian & $92.7 \pm.4$ & $90.9 \pm.4$ & $92.9 \pm.3$ & $88.9 \pm.1$ & $95.1 \pm.2$ & $87.2 \pm.2$ & $87.6 \pm.1$ & $86.8 \pm.2$ & $85.2 \pm.3$ & $86.5 \pm.1$ & $77.9 \pm.2$ & $94.6 \pm.1$ & $85.5 \pm.1$ & $86.0 \pm.5$ & $85.3 \pm.1$ \\
SGD + Impulse & $92.0 \pm.1$ & $90.0 \pm.1$ & $92.3 \pm.1$ & $87.9 \pm.2$ & $94.3 \pm.2$ & $86.1 \pm.2$ & $86.5 \pm.2$ & $85.9 \pm.5$ & $84.2 \pm.2$ & $85.9 \pm.3$ & $76.5 \pm.3$ & $94.3 \pm.2$ & $84.4 \pm.2$ & $84.0 \pm.4$ & $84.5 \pm.3$ \\
\addlinespace
Cutout & $92.8 \pm.2$ & $91.2 \pm.4$ & $93.5 \pm.2$ & $88.2 \pm.9$ & $94.8 \pm.2$ & $86.5 \pm.9$ & $86.9 \pm1$ & $87.0 \pm.6$ & $86.0 \pm.4$ & $85.9 \pm.7$ & $78.0 \pm1$ & $94.8 \pm.2$ & $84.6 \pm.7$ & $84.8 \pm.2$ & $84.4 \pm.5$ \\
Cutout + Gaussian & $92.2 \pm.2$ & $90.2 \pm.3$ & $92.7 \pm.3$ & $87.5 \pm.5$ & $94.3 \pm.2$ & $85.6 \pm.3$ & $85.8 \pm.4$ & $85.7 \pm.4$ & $84.2 \pm.5$ & $84.5 \pm.3$ & $76.3 \pm.5$ & $93.8 \pm.4$ & $83.5 \pm.2$ & $83.7 \pm.3$ & $82.7 \pm.4$ \\
Cutout + Impulse & $91.7 \pm.1$ & $89.6 \pm.1$ & $92.1 \pm.4$ & $87.1 \pm.4$ & $94.2 \pm.4$ & $85.6 \pm.7$ & $85.8 \pm.8$ & $85.8 \pm.6$ & $84.0 \pm.7$ & $85.6 \pm.7$ & $76.4 \pm.7$ & $94.4 \pm.6$ & $84.0 \pm.9$ & $83.4 \pm.6$ & $84.0 \pm.5$ \\
\addlinespace
SAM & $93.3 \pm.2$ & $91.7 \pm.2$ & $93.9 \pm.8$ & $89.5 \pm.5$ & $95.3 \pm.3$ & $87.7 \pm.4$ & $88.2 \pm.5$ & $87.3 \pm.6$ & $85.9 \pm.5$ & $86.6 \pm.5$ & $78.1 \pm.5$ & $94.9 \pm.3$ & $86.1 \pm.3$ & $86.2 \pm.3$ & $86.1 \pm.5$ \\
SAM + Gaussian & $92.5 \pm.7$ & $90.6 \pm.7$ & $92.3 \pm.2$ & $88.2 \pm.4$ & $94.6 \pm.2$ & $86.6 \pm.4$ & $86.7 \pm.5$ & $86.2 \pm.5$ & $84.3 \pm.4$ & $85.7 \pm.2$ & $77.0 \pm.1$ & $94.1 \pm.2$ & $84.8 \pm.3$ & $85.3 \pm.3$ & $84.5 \pm.4$ \\
SAM + Impulse & $91.5 \pm.4$ & $89.6 \pm.5$ & $91.5 \pm.4$ & $86.8 \pm.9$ & $93.7 \pm.6$ & $85.1 \pm.9$ & $85.4 \pm1.0$ & $85.2 \pm.8$ & $83.2 \pm.9$ & $85.0 \pm1.0$ & $75.1 \pm.9$ & $93.6 \pm.6$ & $83.4 \pm.9$ & $83.0 \pm.8$ & $83.5 \pm1.2$ \\
\addlinespace
VIPAug & $94.5 \pm.1$ & $93.0 \pm.1$ & $94.7 \pm.4$ & $90.2 \pm.6$ & $95.5 \pm.1$ & $88.7 \pm.4$ & $89.2 \pm.9$ & $88.7 \pm.3$ & $88.0 \pm.4$ & $87.3 \pm.7$ & $80.9 \pm.5$ & $95.2 \pm.3$ & $86.6 \pm.7$ & $86.8 \pm.2$ & $85.9 \pm.6$ \\
VIPAug + Gaussian & $94.4 \pm.3$ & $92.9 \pm.3$ & $94.5 \pm.1$ & $90.4 \pm.9$ & $95.6 \pm.4$ & $88.9 \pm.7$ & $89.3 \pm.8$ & $88.2 \pm.2$ & $87.7 \pm.5$ & $87.3 \pm.8$ & $80.5 \pm.5$ & $95.2 \pm.3$ & $86.3 \pm.7$ & $86.9 \pm.6$ & $85.6 \pm.6$ \\
VIPAug + Impulse & $93.1 \pm.4$ & $91.1 \pm.4$ & $93.3 \pm.4$ & $88.4 \pm.5$ & $94.2 \pm.1$ & $86.9 \pm.2$ & $87.4 \pm.5$ & $86.6 \pm.2$ & $85.7 \pm.5$ & $86.2 \pm.1$ & $78.5 \pm.5$ & $94.3 \pm.4$ & $84.3 \pm.3$ & $84.2 \pm.2$ & $84.3 \pm.3$ \\
\addlinespace
AugMix & $89.7 \pm.1$ & $87.3 \pm.1$ & $91.7 \pm.2$ & $82.2 \pm.4$ & $91.7 \pm.1$ & $79.4 \pm.3$ & $80.8 \pm.6$ & $82.5 \pm.5$ & $80.7 \pm.3$ & $80.5 \pm.4$ & $72.6 \pm.2$ & $87.6 \pm.3$ & $79.6 \pm.3$ & $81.5 \pm.2$ & $80.4 \pm.3$ \\
AugMix + Gaussian & $89.3 \pm.1$ & $86.8 \pm.1$ & $91.2 \pm.2$ & $80.9 \pm.5$ & $91.3 \pm.2$ & $78.1 \pm.5$ & $79.5 \pm.6$ & $82.0 \pm.4$ & $80.1 \pm.1$ & $79.6 \pm.2$ & $71.4 \pm.2$ & $87.2 \pm.1$ & $78.6 \pm.1$ & $80.4 \pm.1$ & $79.1 \pm.1$ \\
AugMix + Impulse & $89.0 \pm.2$ & $86.5 \pm.2$ & $91.2 \pm.3$ & $80.9 \pm.3$ & $90.7 \pm.1$ & $77.9 \pm.4$ & $79.5 \pm.2$ & $81.3 \pm.1$ & $79.6 \pm.1$ & $79.3 \pm.2$ & $71.4 \pm.2$ & $86.8 \pm.4$ & $78.2 \pm.2$ & $79.6 \pm.2$ & $79.0 \pm.2$ \\
\midrule
\multicolumn{16}{c}{\textbf{ViT}} \\
\midrule
AdamW & $80.2 \pm.4$ & $75.6 \pm.7$ & $79.7 \pm1.2$ & $68.0 \pm.9$ & $81.3 \pm.3$ & $64.4 \pm1.0$ & $66.8 \pm1.4$ & $68.1 \pm.7$ & $63.7 \pm.7$ & $70.7 \pm.8$ & $55.8 \pm.5$ & $81.8 \pm.2$ & $62.9 \pm.6$ & $61.3 \pm.9$ & $61.4 \pm.7$ \\
AdamW + Gaussian & $75.2 \pm.6$ & $70.8 \pm.7$ & $74.4 \pm.9$ & $65.2 \pm.5$ & $79.1 \pm.5$ & $61.6 \pm.5$ & $64.3 \pm.4$ & $65.3 \pm.3$ & $60.9 \pm.3$ & $67.3 \pm.6$ & $52.5 \pm.2$ & $79.3 \pm.1$ & $59.9 \pm.3$ & $58.1 \pm.2$ & $58.4 \pm.5$ \\
AdamW + Impulse & $76.7 \pm.4$ & $72.1 \pm.3$ & $77.5 \pm.8$ & $66.9 \pm.9$ & $77.9 \pm.6$ & $64.1 \pm.9$ & $65.7 \pm1.0$ & $66.4 \pm.9$ & $61.5 \pm.9$ & $69.4 \pm.6$ & $55.1 \pm1.3$ & $79.8 \pm.6$ & $62.2 \pm.9$ & $58.9 \pm1.0$ & $60.2 \pm.9$ \\
\addlinespace
Cutout & $79.8 \pm.4$ & $75.1 \pm.6$ & $79.1 \pm.7$ & $69.6 \pm.8$ & $80.9 \pm.8$ & $64.8 \pm.9$ & $67.4 \pm.9$ & $67.7 \pm.5$ & $63.6 \pm.6$ & $70.7 \pm.6$ & $55.5 \pm1.0$ & $81.7 \pm.4$ & $63.0 \pm.8$ & $61.7 \pm.5$ & $60.8 \pm.9$ \\
Cutout + Gaussian & $74.7 \pm.9$ & $69.7 \pm.8$ & $75.2 \pm1.0$ & $66.0 \pm.8$ & $79.1 \pm.7$ & $62.3 \pm.6$ & $64.7 \pm.7$ & $64.6 \pm.5$ & $60.4 \pm.2$ & $67.1 \pm.2$ & $53.0 \pm.4$ & $79.0 \pm.4$ & $60.5 \pm.4$ & $58.9 \pm.6$ & $59.3 \pm.4$ \\
Cutout + Impulse & $78.2 \pm.9$ & $73.8 \pm1.1$ & $79.7 \pm.4$ & $67.4 \pm.6$ & $79.1 \pm1.3$ & $64.2 \pm.9$ & $66.9 \pm.5$ & $67.4 \pm.8$ & $62.2 \pm.9$ & $70.7 \pm.8$ & $55.9 \pm.4$ & $81.8 \pm.7$ & $62.9 \pm.5$ & $61.1 \pm1.3$ & $60.8 \pm.6$ \\
\addlinespace
SAM & $72.2 \pm.1$  & $67.3 \pm.1$  & $70.7 \pm.2$  & $65.0 \pm.1$  & $78.2 \pm.1$  & $58.3 \pm.2$  & $63.6 \pm.2$  & $64.1 \pm.1$  & $59.6 \pm.2$  & $64.8 \pm.2$  & $51.1 \pm.2$  & $78.4 \pm.1$  & $59.8 \pm.1$  & $52.7 \pm.2$  & $52.8 \pm.2$  \\
SAM + Gaussian & $69.5 \pm.4$  & $64.7 \pm.3$  & $68.8 \pm.3$  & $64.5 \pm.4$  & $77.5 \pm.1$  & $57.9 \pm.5$  & $62.9 \pm.6$  & $63.3 \pm.1$  & $59.1 \pm.2$  & $64.9 \pm.1$  & $51.1 \pm.5$  & $78.2 \pm.4$  & $59.5 \pm.5$  & $52.5 \pm.1$  & $52.6 \pm.1$  \\
SAM + Impulse & $69.9 \pm.6$  & $64.9 \pm.8$  & $68.6 \pm.8$  & $64.5 \pm.4$  & $74.2 \pm1.0$  & $58.1 \pm.8$  & $62.3 \pm.9$  & $63.8 \pm.3$  & $58.5 \pm.4$  & $66.0 \pm.2$  & $52.8 \pm.5$  & $78.7 \pm.7$  & $59.6 \pm.6$  & $55.3 \pm.6$  & $54.6 \pm.6$  \\
\addlinespace
VIPAug & $80.0 \pm.6$  & $75.4 \pm.5$  & $80.2 \pm.5$  & $68.9 \pm.8$  & $81.4 \pm.1$  & $64.9 \pm.7$  & $67.5 \pm.6$  & $66.7 \pm.6$  & $63.0 \pm.2$  & $69.9 \pm.5$  & $54.8 \pm.7$  & $82.1 \pm.2$  & $62.7 \pm.7$  & $61.6 \pm.4$  & $60.9 \pm.6$  \\
VIPAug + Gaussian & $75.3 \pm.6$  & $70.9 \pm.7$  & $74.8 \pm1.0$  & $65.3 \pm.2$  & $79.5 \pm.6$  & $61.5 \pm.3$  & $64.2 \pm.3$  & $64.8 \pm.4$  & $61.0 \pm.4$  & $67.3 \pm.6$  & $51.9 \pm.1$  & $78.9 \pm.3$  & $60.0 \pm.1$  & $59.2 \pm.5$  & $59.0 \pm.2$  \\
VIPAug + Impulse & $76.6 \pm1.6$  & $72.3 \pm1.6$  & $78.5 \pm1.5$  & $66.8 \pm.8$  & $78.7 \pm.7$  & $63.4 \pm.8$  & $65.8 \pm.5$  & $65.5 \pm1.4$  & $61.3 \pm.9$  & $69.0 \pm.8$  & $54.7 \pm.6$  & $81.0 \pm.6$  & $61.2 \pm.8$  & $60.2 \pm.4$  & $59.9 \pm.7$  \\
\addlinespace
AugMix & $71.1 \pm.1$  & $65.8 \pm.1$  & $70.3 \pm.2$  & $61.0 \pm.3$  & $76.0 \pm.1$  & $55.5 \pm.3$  & $59.6 \pm.2$  & $62.1 \pm.3$  & $59.1 \pm.1$  & $61.8 \pm.3$  & $50.3 \pm.2$  & $73.3 \pm.1$  & $57.9 \pm.2$  & $55.7 \pm.1$  & $56.2 \pm.2$  \\
AugMix + Gaussian & $66.5 \pm.8$  & $62.0 \pm.9$  & $66.2 \pm.9$  & $60.5 \pm.1$  & $74.4 \pm.5$  & $54.8 \pm.1$  & $58.7 \pm.1$  & $61.2 \pm.3$  & $57.8 \pm.2$  & $60.8 \pm.1$  & $49.8 \pm.1$  & $72.4 \pm.1$  & $57.4 \pm.1$  & $54.2 \pm.6$  & $55.0 \pm.6$  \\
AugMix + Impulse & $69.4 \pm.1$  & $64.2 \pm.1$  & $69.0 \pm.1$  & $60.3 \pm.1$  & $75.2 \pm.2$  & $55.0 \pm.1$  & $59.0 \pm.2$  & $61.5 \pm.1$  & $58.3 \pm.1$  & $61.1 \pm.1$  & $49.8 \pm.1$  & $73.0 \pm.1$  & $57.3 \pm.1$  & $55.3 \pm.1$  & $55.6 \pm.2$  \\
\bottomrule
\end{tabular}
}
\end{sidewaystable}

\section{Noise period ablation results}
\autoref{tab:ablation_P} shows an ablation over the noise injection period P run on the CIFAR-100 dataset with the WideResNet model. For the results, we chose P based on the strongest performance on the clean data, resulting in a period of 5 used for all trials (we note P=10 also performs strong for the Gaussian noise, but is worse for impulse noise).

\begin{table}[h]
\centering
\caption{Ablation over the interleaved period P showing the same performance metrics with the same model/dataset as \autoref{tab:ablation_f}. Note $n_{\rm{inj}}$ denotes the total amount of noise injections during the 100 epoch training run.}
\label{tab:ablation_P}
\begin{tabular}{c c ccc ccc}
\toprule
 &  & \multicolumn{3}{c}{\textbf{Gaussian noise}} & \multicolumn{3}{c}{\textbf{Impulse noise}} \\
\cmidrule(lr){3-5} \cmidrule(lr){6-8}
period (P) & \textbf{$n_{\rm{inj}}$}\ \ \ & Clean & mCE & Struct. & Clean & mCE & Struct. \\
\midrule
0 & 100 & $43.4 \pm.1$ & $50.7 \pm.1$ & $51.9 \pm.1$ & $42.0 \pm.1$ & $56.0 \pm.2$ & $50.0 \pm.1$ \\
1 & 50 & $28.3 \pm.2$ & $\mathbf{42.3 \pm.2}$ & $\mathbf{41.5 \pm.2}$ & $25.3 \pm.2$ & $\mathbf{48.8 \pm.2}$ & $\mathbf{42.3 \pm.2}$ \\
2 & 33 & $26.0 \pm.1$ & $43.9 \pm.2$ & $41.7 \pm.1$ & $25.3 \pm.1$ & $49.0 \pm.1$ & $42.5 \pm.2$ \\
5 & 20 & $25.0 \pm.1$ & $46.3 \pm.1$ & $42.8 \pm.1$ & $\mathbf{24.9 \pm.2}$ &$50.4 \pm.1$ & $44.3 \pm.1$ \\
10 & 10 & $\mathbf{24.9 \pm.1}$ & $47.5 \pm.1$ & $43.2 \pm.1$ & $25.4 \pm.1$ & $51.3 \pm.1$ & $45.3 \pm.1$ \\
20 & 5 & $25.5 \pm.2$ & $50.6 \pm.1$ & $45.5 \pm.2$ & $25.5 \pm.1$ &$51.8 \pm.2$ & $46.0 \pm.1$ \\ 
40 & 2 & $25.5 \pm.1$ & $50.2 \pm.1$ & $45.1 \pm.1$ & $25.7 \pm.2$ &$51.8 \pm.1$ & $46.2 \pm.1$ \\ 
50 & 1 & $25.6 \pm.1$ & $52.2 \pm.2$ & $46.7 \pm.2$ & $25.6 \pm.1$ &$52.8 \pm.1$ & $47.0 \pm.2$ \\ \midrule
\rowcolor{red!8} no noise injection & & $25.7 \pm.1$ & $52.6 \pm.1$ & $47.1 \pm.4$ & $25.7 \pm.3$ &$52.6 \pm.2$ & $47.1 \pm.2$ \\
\bottomrule
\end{tabular}
\end{table}

\section{Noise corruption visualization}
\begin{figure}[h]
    \centering
    \includegraphics[width=0.95\linewidth]{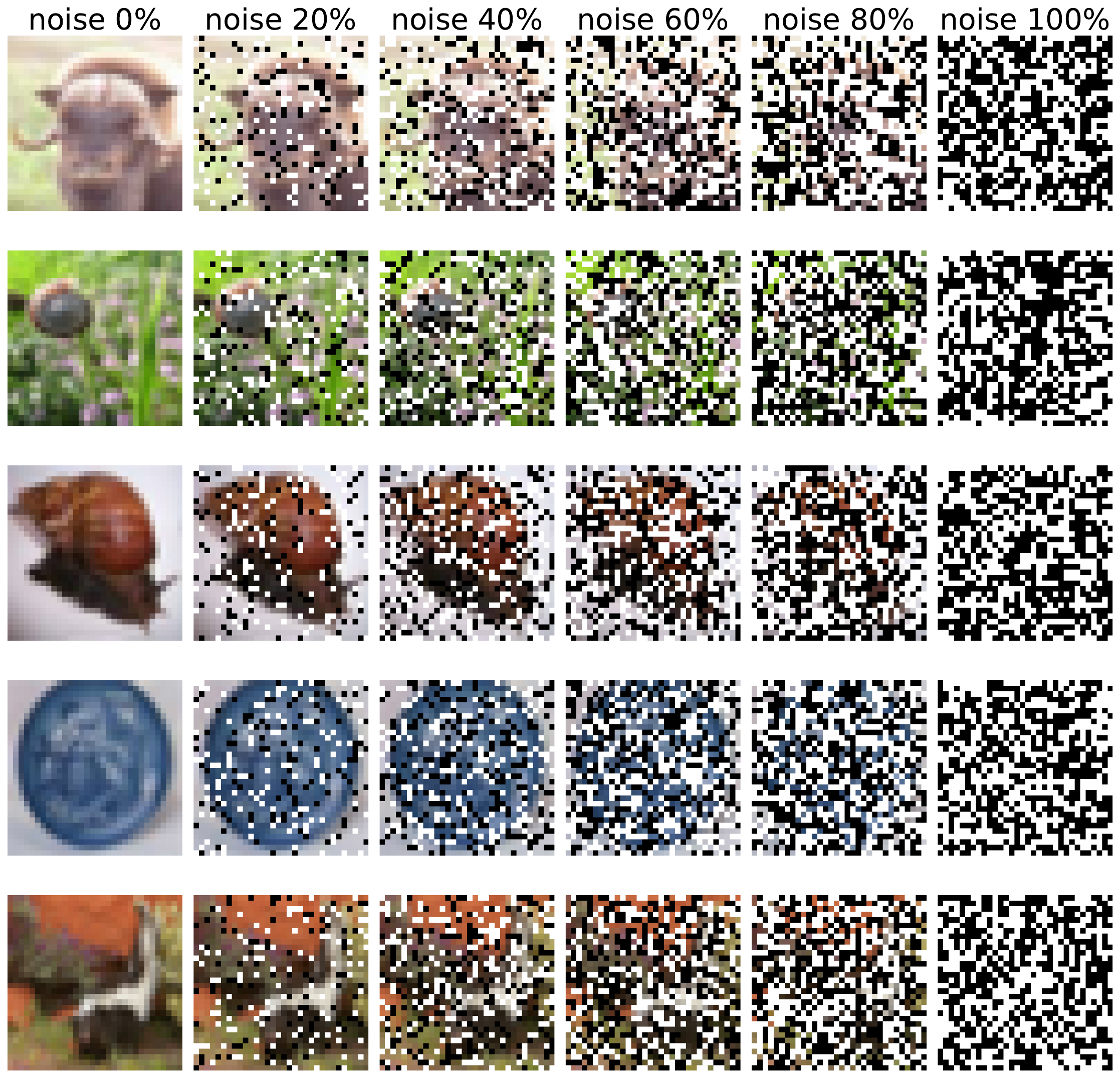}
    \caption{CIFAR-100 samples with increasing levels of impulse noise, with the percentages corresponding to the probability ($\sigma$) of individual pixel corruption.}
    \label{fig:noise-grid}
\end{figure}

We show a visualisation of different impulse noise levels on the CIFAR-100 dataset in \autoref{fig:noise-grid}.

\end{document}